\documentclass[10pt,twocolumn,letterpaper]{article}

\usepackage{iccv}
\usepackage{times}
\usepackage{epsfig}
\usepackage{graphicx}
\usepackage{amsmath}
\usepackage{amssymb}
\usepackage[font={small}]{caption}
\usepackage{enumitem}
\usepackage{booktabs}
\usepackage{color}
\usepackage{xcolor}
\usepackage{tabulary,multirow,overpic,xcolor}
\usepackage{verbatim}
\usepackage{float}
\usepackage{nicefrac}       %
\usepackage{microtype}      %
\usepackage[caption=false]{subfig}
\usepackage{makecell}
\def\x{$\times$}
\newcolumntype{x}[1]{>{\centering\arraybackslash}p{#1pt}}

\usepackage[pagebackref=true,breaklinks=true,letterpaper=true,colorlinks,citecolor=citecolor,bookmarks=false]{hyperref}

\iccvfinalcopy %
\usepackage{color}
\usepackage{xcolor}
\definecolor{citecolor}{HTML}{0071bc}
\usepackage[pagebackref=true,breaklinks=true,colorlinks,citecolor=citecolor,bookmarks=false]{hyperref}

\newlength\savewidth\newcommand\shline{\noalign{\global\savewidth\arrayrulewidth
  \global\arrayrulewidth 1pt}\hline\noalign{\global\arrayrulewidth\savewidth}}
\newcommand{\tablestyle}[2]{\setlength{\tabcolsep}{#1}\renewcommand{\arraystretch}{#2}\centering\footnotesize}

\def\iccvPaperID{1710} %

\ificcvfinal\fi

\begin{document}
\title{Video Autoencoder: self-supervised disentanglement\\  of static 3D structure and motion}

\author{Zihang Lai\\
Carnegie Mellon University\\
\and
Sifei Liu\\
NVIDIA\\
\and
Alexei A. Efros\\
UC Berkeley\\
\and
Xiaolong Wang\\
UC San Diego\\
}

\twocolumn[{%
\vspace{-1em}
\maketitle
\vspace{-1em}

\begin{center}
    \centering 

    \vspace{-0.3in}
    \includegraphics[width=0.95\textwidth]{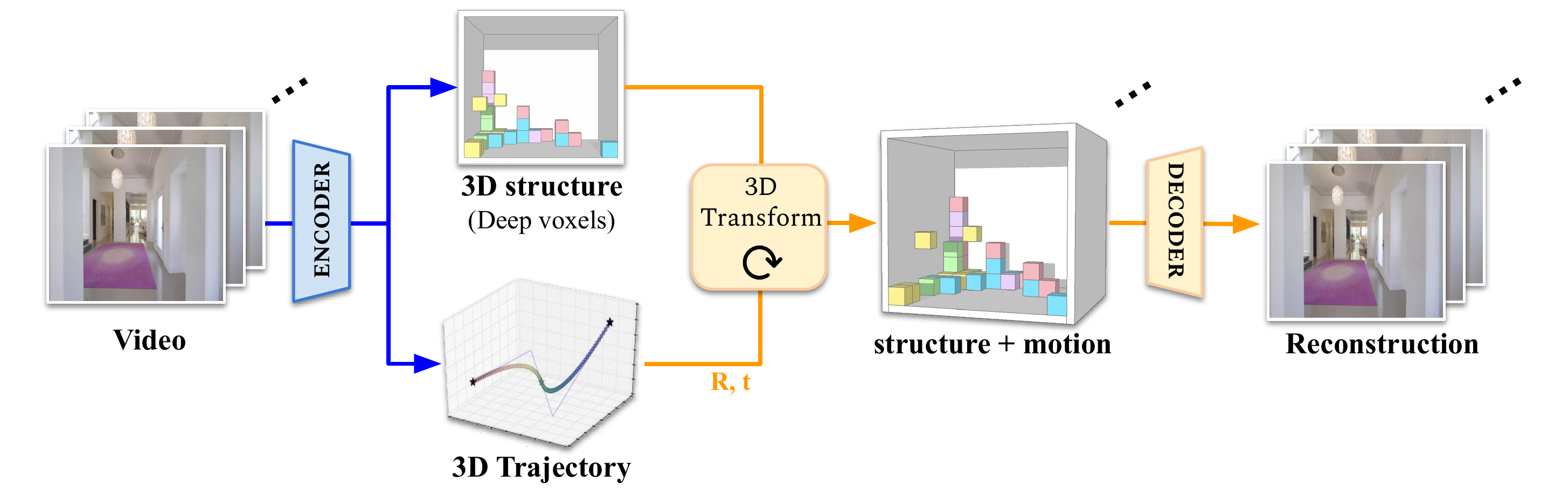}
    \vspace{-0.15in}
    \captionof{figure}{\textbf{Video Autoencoder:} 
      raw input video is automatically disentangled into 3D scene structure and camera trajectory. To reconstruct the original video, the camera transformation is applied to the 3D structure feature and then decoded back to pixels.  Without any fine-tuning, the model generalizes to unseen videos and enables tasks such as novel view synthesis, pose estimation, and ``video following''. }
    \label{teaser}
\end{center}
}] 
\ificcvfinal\thispagestyle{empty}\fi

\begin{abstract}
\vspace{-0.05in}
A video autoencoder is proposed for learning disentangled representations of 3D structure and camera pose from videos in a self-supervised manner. Relying on temporal continuity in videos, our work assumes that the 3D scene structure in nearby video frames remains static. Given a sequence of video frames as input, the video autoencoder extracts a disentangled representation of the scene including: (i) a temporally-consistent deep voxel feature to represent the 3D structure and (ii) a 3D trajectory of camera pose for each frame. These two representations will then be re-entangled for rendering the input video frames. This video autoencoder can be trained directly using a pixel reconstruction loss, without any ground truth 3D or camera pose annotations. The disentangled representation can be applied to a range of tasks, including novel view synthesis, camera pose estimation, and video generation by motion following. We evaluate our method on several large-scale natural video datasets, and show generalization results on out-of-domain images. Project page with code:  \href{https://zlai0.github.io/VideoAutoencoder}{https://zlai0.github.io/VideoAutoencoder}.
\vspace{-0.05in}
\end{abstract}

\section{Introduction}

The visual world arrives at a human eye as a streaming, entangled mess of colors and patterns. The art of seeing, to a large extent, is in our ability to disentangle this mess into physically and geometrically coherent factors: persistent solid structures, illumination, texture, movement, change of viewpoint, etc.  From its very beginnings, computer vision has been concerned with acquiring this impressive human ability, including 
such classics as Barrow and Tenebaum's Intrinsic Image decomposition~\cite{barrow1978recovering} in the 1970s, or Tomasi-Kanade factorization~\cite{tomasi1992shape} in the 1990s.  In the modern deep era, learning a disentangled visual representation has been a hot topic of research, often taking the form of an  autoencoder
~\cite{kulkarni2015deep,higgins2016beta,kim2018disentangling,harsh2018disentangling,liu2020factorize,park2020swapping,heljakka2020deep,anokhin2020high,pidhorskyi2020adversarial}.  However, almost all prior work has focused on disentanglement within the 2D image plane using datasets of still images. 

In this work, we propose a method that learns a disentangled 3D scene representation, separating the static 3D scene structure from the camera motion.   
Importantly, we employ videos as training data (as opposed to a dataset of stills), using the temporal continuity within a video as a source of training signal for self-supervised disentanglement. 
We make the assumption that a local snippet of video is capturing a static scene, so the changes in appearance must be due to camera motion.
This leads to our {\em Video Autoencoder} formulation, shown on Figure~\ref{teaser}:  an input video is encoded into two codes, one for 3D scene structure (which is forced to remain fixed cross frames) and the other for the camera trajectory (updated for every frame). The 3D structure is represented by 3D deep voxels (similar to~\cite{nguyen2019hologan,sitzmann2019deepvoxels}) and the camera pose with a 6-dimension rotation and translation vector. 
To reconstruct the original video, we simply apply the camera transformation to the 3D structure features and then decode back to pixels.

A key advantage of our framework is that it provides 3D representations readily integrated modern neural rendering methods, which typically requires 3D and/or camera pose ground truth annotation at training time~\cite{flynn2016deepstereo,niklaus20193d,bansal20204d,wiles2020synsin, tewari2020state}. 
This usually implies a 2-stage process with running Structure-from-Motion (SfM) as precursor to training. In our work, we are working towards a way of training on completely unstructured datasets, removing the need for running SfM as preprocessing.

At test time, the features obtained using our Video Autoencoder can be used for several downstream tasks, including novel view synthesis (Section~\ref{sec:viewsynthesis}), pose estimation in video (Section~\ref{sec:pose}), and video following (Section~\ref{sec:videofollowing}). For novel view synthesis, given a single input image, we first encode it 
as a 3D scene feature, and then render to a novel view by providing a new camera pose.  We show results on large-scale video datasets including RealEstate10K~\cite{zhou2018stereo}, Matterport3D~\cite{Matterport3D}, and Replica~\cite{replica19arxiv}. Our method not only achieves better view synthesis results than state-of-the-art view synthesis approach~\cite{wiles2020synsin} that requires stronger camera supervision on RealEstate10K, but also generalizes better when applied to out-of-domain data.  As another application, we show that our method could be used to implicitly factorize structure from motion in novel videos, by evaluating the estimated camera pose against SfM baseline.
Finally, we show that by swapping the 3D structure and camera trajectory codes between a pair of videos, we can achieve Video Following, where a scene from one video is ``following'' the motion from the other video.

\section{Related Work}

\textbf{Learning disentangled representations}. Disentangled representations learned from unlabeled data not only provide a better understanding of the data, but also produce more generalizable features for different downstream applications. Popular ways to learn such representations include generative models (e.g. GANs)~\cite{chen2016infogan,Wang_ssganECCV2016,VON,huang2018munit,karras2019style,nguyen2019hologan,shen2020interpreting,peebles2020hessian,lee2020drit} and autoencoders~\cite{kulkarni2015deep,higgins2016beta,kim2018disentangling,harsh2018disentangling,liu2020factorize,park2020swapping,heljakka2020deep,anokhin2020high,pidhorskyi2020adversarial}. For example, Kulkarni et al.~\cite{kulkarni2015deep} proposed to learn disentangled representations of pose, light, and shape for human faces using a Variational Autoencoder (VAE)~\cite{kingma2013auto}. However, almost all these works are modeling still 2D images, inherently limiting the data available for disentanglement. 
In order to learn disentangled representations related to motion and dynamics, researchers have been looking at video data
~\cite{denton2017unsupervised,YeTian_physicsECCV2018,jakab2018unsupervised,Wiles18a,tulyakov2018mocogan,hsieh2018learning,minderer2019unsupervised,xue2016visual}. For example, Denton et al.~\cite{denton2017unsupervised} proposed to learn the disentangled representation which factorizes each video frame into a stationary component and a temporally varying component. Beyond learning latent features, Tomas et al.~\cite{jakab2018unsupervised} designed a video frame reconstruction method for disentangling human pose skeleton from frame appearance. While these results are encouraging, they are not able to capture the 3D structure of generic scenes from video.

\textbf{Learning 3D representations.} 3D representation learning from video or 2D image sets is a long-standing problem in computer vision. Traditional approaches typically rely on multi-view geometry~\cite{hartley2003multiple} to understand real-world 3D structures. Based on geometric principles, 3D structure and camera motion can be jointly optimized in Structure-from-Motion (SfM) pipelines and has yielded great success in a wide range of domains~\cite{agarwal2009building, schoenberger2016sfm, snavely2008scene}. To better generalize to diverse environments, learning-based approaches are proposed to learn 3D representations using 2D supervision~\cite{jimenez2016unsupervised,zhou2017unsupervised,tulsiani2017multi,kanazawa2018learning,wiles2020synsin,hu2021worldsheet,Rockwell2021}. For instance, Wiles et al.~\cite{wiles2020synsin} proposed to utilize the point cloud as an intermediate representation for novel view synthesis. However, it requires camera poses computed from SfM for training, and point cloud estimation can be inaccurate when the test image is out of distribution. Instead of using point clouds, neural 3D representations, including implicit function~\cite{mildenhall2020nerf,sitzmann2019scene} and the deep voxels~\cite{Tung_2019_CVPR,harley_viewcontrast,sitzmann2019deepvoxels,nguyen2019hologan,mustikovela2020self} have shown impressive reconstruction and synthesis results.  Our work is closely related to the approach proposed by Tung et al.~\cite{Tung_2019_CVPR}, which leverages view prediction for learning latent 3D voxel structure of the scene. However, camera pose is still required to provide supervision. Our work is also highly inspired by  Nguyen-Phuoc et al.~\cite{nguyen2019hologan}, who proposed inserting the voxel representation into Generative Adversarial Networks, enabling the disentanglement of 3D structure, style, and pose. Finally, our work is related to plenty of downstream tasks that leverages a learned 3D deep voxels, such as 3D object detection~\cite{Tung_2019_CVPR}, 3D object tracking~\cite{harley2020tracking}, 3D motion estimation~\cite{harley_viewcontrast} and few-shot concept learning~\cite{prabhudesai2021disentangling}.

\textbf{Self-supervised learning on video.} Our work is related to self-supervised learning of visual representations from video~\cite{agrawal2015learning,Wang15,Isola15,Jayaraman15,Jayaraman16,Misra16,zhou2017unsupervised,Wiles18a,Wei18,Tung_2019_CVPR,Han19}. For example, Wei et al.~\cite{Wei18} proposed to learn from the arrow of time and obtain a representation that is sensitive to temporal changes; it can then be used for action recognition. Instead of fine-tuning the learned representation for recognition, our work is more focused on the 3D structure of the representation itself, and we can directly adopt our representation for multiple applications without fine-tuning. In this regard, our work is more related to Zhou et al.~\cite{zhou2017unsupervised}, who perform joint estimation of image depth and camera pose in a self-supervised manner. However, their approach is restricted to specific domains, such as scenes from self-driving cars, while our model allows generalization to a wider range of real-world videos. Instead of predicting depth, our method uses a voxel representation, which  can be applied to downstream tasks, such as novel view synthesis.

\begin{figure*}[t]
    \centering
    \includegraphics[width=\textwidth]{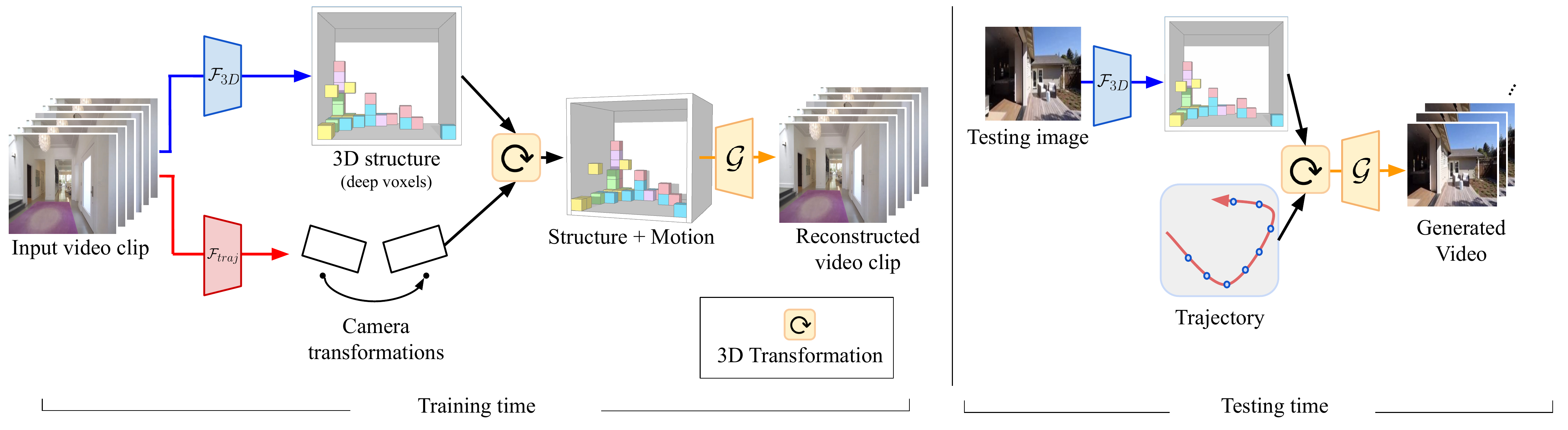}
    \vspace{-0.3in}
    \caption{\textbf{Training and testing procedure of Video Autoencoder:} During training, we use short clips extracted from videos. The first frame of the clip is used to predict the 3D structure of the scene. Then, the subsequent frames are used to compute poses relative to the first frame. We apply these predicted poses as affine transformations to the 3D voxel and use a decoder to reconstruct the input video clip. Once the autoencoder is trained, we can get the 3D representation for downstream tasks using only one image as input.}
    \label{fig:structure}
\end{figure*}
\section{Approach}

The proposed Video Autoencoder is a conceptually simple method for encoding a video into a 3D representation and a trajectory in a completely self-supervised manner (\textbf{no 3D labels} are required). 
Figure~\ref{fig:structure} shows a schematic layout of our Video Autoencoder. 
Like other auto-encoders, we \textit{encode} data into a deep representation and \textit{decode} the representation back to reconstruct the original input, relying on the consistency between the input and the reconstruction to learn sub-modules of multiple neural networks. 
In our case, the goal is encoding a video into two \textit{disentangled} components: a static 3D representation and a dynamic camera trajectory.
By assuming that the input video clip shows a static scene that remains unchanged in the video clip, we can construct a single 3D structure (represented by deep voxels) and apply camera transformations on the structure to reconstruct corresponding video frames. Unlike other existing methods~\cite{wiles2020synsin, single_view_mpi, mildenhall2020nerf, zhou2016view}, our model does not need ground truth camera poses. Instead, we use another network to predict the camera motion, which is then jointly optimized with the 3D structure. 
By doing so, we find that the 3D motion and structure can automatically emerge from the auto-encoding process.

\textbf{Training.} At training time, the \emph{first frame} of an $N$-frame video clip (we use $N=6$) passes through the 3D Encoder (\textit{blue box} in Fig.~\ref{fig:structure}), which computes a voxel grid of deep features representing the 3D scene.  At the same time, the trajectory encoder (\textit{red box} in Fig.~\ref{fig:structure}) uses the same video clip to produce a short trajectory of three points. This trajectory estimates the camera pose of each frame w.r.t. the first frame. Next, we \textit{re-entangle} the camera trajectory and the 3D structure and reconstruct the input video clip. First, we use the estimated camera pose to transform the encoded 3D deep voxel. Because we assume that the scene is static, the transformed 3D deep voxel should align with the corresponding frame if both the voxel representation and the camera pose are accurately estimated.
We then use a decoder network to render $N$ 2D images from the set of $N$ camera pose transformed 3D voxels. 
The reconstruction loss encourages the disentanglement between the static 3D scene and the camera pose. We also adopt a consistency loss to enforce the 3D deep voxels extracted from different frames to be the same, which facilitates training.

\textbf{Inference for View Synthesis.} The procedure during test time is similar. First, the 3D Encoder estimates the 3D voxel representation from a single input image. The trajectory can be of arbitrary length and pre-computed. Next, we compute the transformed 3D deep voxels according to the given camera trajectory. These trajectory-guided deep voxels are then fed into the decoder which renders each frame of the video as outputs.
While this approach works well for nearby frames where the motion is not too large, we found that voxels can fail to render clear images if the applied transformation is too large. Therefore, when we need to generate long videos at test time, we employ a simple heuristic which \textit{reinitializes} the deep voxels to the current frame every $K$ frames ($K=12$ in our implementation). Here, the \textit{reinitialize} operation re-encodes the previous prediction into a new 3D voxel to replace the existing 3D voxel.

We describe each individual component of our architecture in the following sections. In section~\ref{ssec:encoder3d}, we give details of how we obtain the 3D latent voxel representation from a single image. In section~\ref{ssec:encodertraj}, we describe the method for predicting trajectories for a particular video. In section~\ref{ssec:decoder}, the decoder which \textit{re-entangles} camera motion and 3D structure back into image space is presented. In section~\ref{ssec:loss}, we discuss the loss function used for training the auto-encoder. 

\subsection{3D Encoder} \label{ssec:encoder3d}
\begin{figure}[t]
    \centering
    \includegraphics[width=\linewidth]{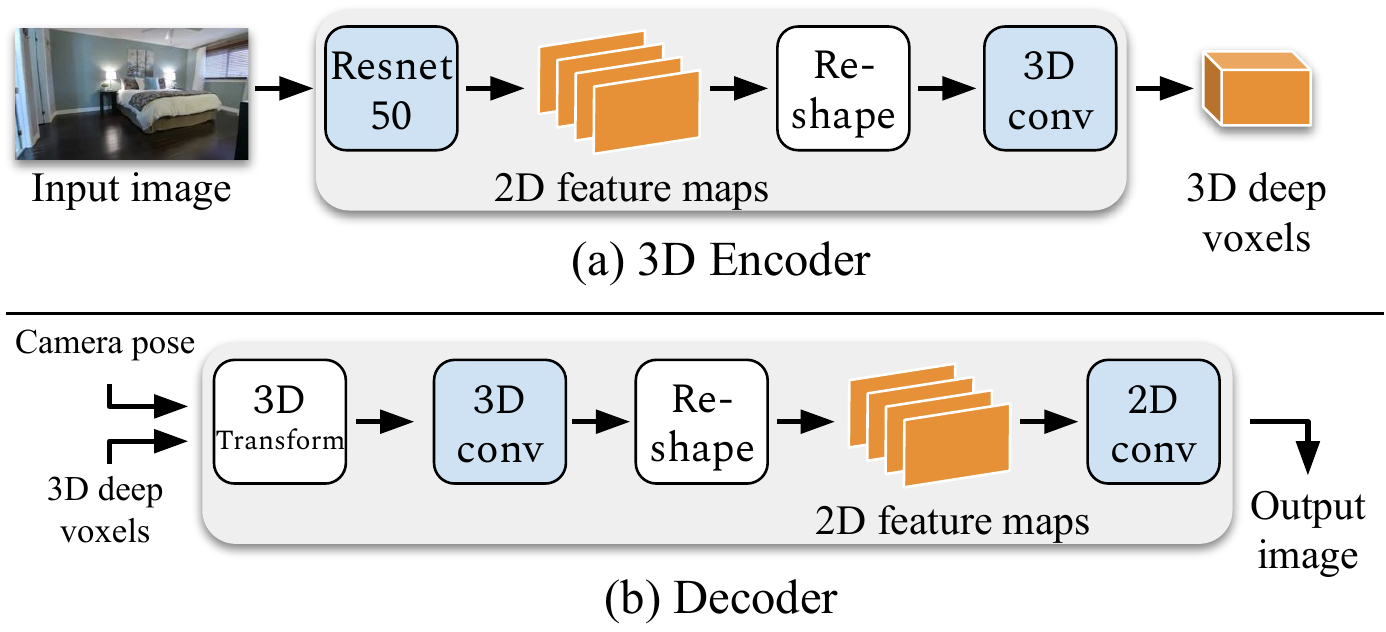}
    \vspace{-0.2in}
    \caption{\textbf{Structure of 3D Encoder and Decoder:} The 3D encoder takes an image as input and estimates a 3D deep voxel corresponding to the same scene. Conversely, the decoder takes the 3D voxel and a camera pose as input and renders a 2D image as output.}
    \label{fig:encoderdecoder}
\end{figure}
The 3D encoder $\mathcal{F}_{\text{3D}}$ encodes an image input into a 3D deep voxels that represents the same scene as the output, 
\[z = \mathcal{F}_{\text{3D}}(I)\]
Figure~\ref{fig:encoderdecoder} (a) illustrates the structure of 3D Encoder in detail. Taking an image as input, we first use a 2D encoder (a pretrained ResNet-50~\cite{he2016deep} in our implementation) to compute a set of 2D feature maps (\textit{Resnet-50} in Figure~\ref{fig:encoderdecoder} (a)). Next, to obtain a 3D representation of the image, we reshape these 2D feature maps into 3D feature grids, which are also referred to as deep voxels. Here, the \textit{reshape} operation is performed on the feature dimension: if the 2D feature maps have dimension $H\times W\times C$, the reshaped tensor is four-dimensional and has size $H\times W\times D \times (C/D)$, where D refers to depth dimension. This ensures that the spatial arrangement is not perturbed, \textit{e.g.} the top right corner of the deep voxels corresponds to the top right corner of the input image. Because the 2D feature extractor repeatedly downsamples the input image, the spatial resolution of the deep voxels is actually small (only $1/16$ of the original image). In order to reconstruct images of high fidelity, we upsample and refine this 3D structure as a final step. This is done by a set of strided 3D deconvolutions (\textit{3D Conv} in Figure~\ref{fig:encoderdecoder} (a)). 

Note while there are other possible 3D representations such as point cloud and polygon mesh, none of these methods are as easy to work with as voxels, which allows reshaping and could be applied with convolutions easily. We use voxel representation to keep things simple, but other representations could potentially work as well.

\subsection{Trajectory Encoder} \label{ssec:encodertraj}
The trajectory encoder $\mathcal{F}_{\text{traj}}$ estimates trajectory from input videos. 
Specifically, the encoder computes the camera pose (6-D, rotation and translation) w.r.t. the first frame for each image in a sequence. 
We call this output sequence of camera poses the \textit{trajectory}. 
Because we compute camera pose for every video frame, the length of the trajectory is the same as the number of frames. 
To compute the relative pose between a particular target image and the reference image (\textit{i.e.,} the first frame), we make use of a simple ConvNet $\mathcal{H}$. 
The network takes as input both the target and reference image, stacked along the channel dimension (\textit{i.e.,} the input channel is 6), and computes a 6-dim vector through a series of seven 2D convolutions. 
We use this vector as the 3D rotation and translation between the reference image and the target image, and it is used for transforming the deep voxels (Sec.~\ref{ssec:decoder}).
Overall, we obtain the trajectory as
\[E = \mathcal{F}_{\text{traj}}(V) = \{\mathcal{H}(I_1, I_i)\}_{i=1}^N\]

\subsection{Decoder} \label{ssec:decoder}
The decoder $\mathcal{G}$ is very similar to an inverse process of the 3D encoder: it renders a 3D deep voxel representation back into image space with a given camera transformation. 
\[\hat{I} = \mathcal{G}(Z, e) \text{,\hspace{0.5em}} e\in E\]

Fig.~\ref{fig:encoderdecoder} (b) illustrates the architecture of the decoder. Taking camera pose and 3D deep voxels as input, the decoder first applies the 3D transformation (\textit{3D Transform} in Figure~\ref{fig:encoderdecoder} (b)) of the camera pose on the deep voxels. If the camera pose and the 3D representation are both correct, the transformed 3D voxels should match the frame corresponds to the camera pose. Specifically, we warp the grid such that the voxel at location $p=(i,j,k)^T$ will be warped to $\hat{p}$, which is computed as 
\[\hat{p} = Rp+t\]
where $R, t$ is the $3\times 3$ rotation matrix and translation vector corresponding to the camera pose. In our implementation, the warp is performed inversely and the value at fractional grid location is trilinearly sampled. Due to the coarseness of the voxel representation, there could be misaligned voxels during the sampling procedure. We thus apply two 3D convolutions to refine and correct these mismatches (\textit{3D Transform} in Figure~\ref{fig:encoderdecoder} (b)). The refined voxels are then reshaped back into the 2D feature maps. To align with the similar \textit{reshape} process in 3D Encoder, we concatenate the feature dimension and the depth dimension. That is, if the input 3D deep voxel has dimension $H\times W\times D\times C$, the reshaped tensor will be of size $H\times W\times (D*C)$. This is a set of 2D feature maps which we then use several layers of 2D convolutions to map them back to an image. The output image has the original resolution. In our implementation, $H=64, W=64, D=32, C=32$.

\subsection{Training Loss} \label{ssec:loss}
We apply a reconstruction loss between reconstructed video clips and the original video clips. The loss is defined as,
\[L_{\text{recon}}(\hat{I_t},I_t) = \lambda_{L1} ||\hat{I_t}-I_t||_1 + \lambda_{\text{perc}} L_{\text{perc}}(\hat{I_t}, I_t) \]
where $I_t$ is the original video frame at time $t$ and $\hat{I_t}$ is the reconstructed frame at time $t$; $L_{\text{perc}}$ denotes the VGG-16 perceptual loss~\cite{johnson2016perceptual,dosovitskiy2016inverting}. In our experiments, $\lambda_{L1} = 10$, $\lambda_{\text{perc}} = 0.1$. To enhance the image quality of reconstructed images, we also apply a WGAN-GP~\cite{gulrajani2017improved} adversarial loss on each output frame in addition to the reconstruction loss. This adversarial loss comes from a separate critic function $\mathcal{F}_D$ which learns to decide if the image looks realistic. Formally, the WGAN-GP minimizes the value function,
\[\min_{G} \max_{\mathcal{F}_D\in \mathcal{D}} 
\mathop{\mathbb{E}}_{I\in \mathbb{P}_r}[\mathcal{F}_D(I)] - 
\mathop{\mathbb{E}}_{\hat{I}\in \mathbb{P}_g}[\mathcal{F}_D(\hat{I})]\]
and thereby minimizes the Wasserstein distance between the data distribution defined by the training set and the model distribution induced by Video Autoencoder.
Here $G$ is our model, $\mathcal{D}$ is the set of 1-Lipschitz functions, $\mathbb{P}_r$ is data distribution and $\mathbb{P}_g$ is the model distribution.
The loss $L_{\text{GAN}}$ is thus defined as the negated critic score for each image:
\[L_{\text{GAN}}(\hat{I}) = -\mathcal{F}_D(\hat{I})\]

Finally, in order to ensure that a \textit{single} 3D structure is used to represent different frames of the same scene, we apply a consistency loss between the 3D voxels extracted from different frames. Specifically, we want to ensure that any pairs of images from the same video, $I_{t1}, I_{t2}$, should be encoded into the same 3D representation, after rotating by the relative camera motion. Formally, we apply consistency loss 
\[L_{\text{cons}}(I_{t1},I_{t2})=||\mathcal{R}(\mathcal{F}_{\text{3D}}(I_{t1}),\mathcal{H}(I_{t1},I_{t2})) - \mathcal{F}_{\text{3D}}(I_{t2})||_1\]
where we enforce that the 3D deep voxel encoded from frame $I_{t1}$, after transformed by the relative pose 
between $I_{t1}$ and $I_{t2}$, should be consistent with the 3D deep voxel encoded from frame $I_{t2}$. $\mathcal{F}_{\text{3D}}$ and $\mathcal{H}$ are described in Sec.~\ref{ssec:encoder3d} and \ref{ssec:encodertraj}, respectively.
$\mathcal{R}$ is the 3D transformation function described in Sec.~\ref{ssec:decoder}.

Overall, our final loss is:
\[L = \sum_t{L_{\text{recon}}(I_t,\hat{I_t}) + \lambda_{\text{GAN}}L_{\text{GAN}}(\hat{I_t}) + \lambda_{\text{cons}}L_{\text{cons}}(I_0,I_t) }\]

In our experiments, we use $\lambda_{\text{cons}} = 1$, $\lambda_{\text{GAN}} = 0.01$.

\section{Experiments}

In this section, we empirically evaluate our method and compare it to existing approaches on three different tasks: camera pose estimation, single image novel view synthesis,  and video following. We show that, although our method is quite simple, it performs surprisingly well against more complex existing methods. 

\subsection{Implementation Details}
As preprocessing, we resize all images into a resolution of $256\times 256$. During training, the training video clip consists of 6 frames. When we train on the RealEstate10K dataset~\cite{zhou2018stereo}, these 6 frames are sampled at a frame-rate of 4 fps so that the motion is sufficiently large. For training on Matterport3D~\cite{Matterport3D}, we do not sample with intervals because the motion between frames is already large. The depth dimension of the 3D deep voxels is set to $D=32$ in our implementation.  We train our model end-to-end using a batch size of 4 for 200K iterations with an Adam optimizer~\cite{kingma2014adam}. The initial learning rate is set to $2e^{-4}$ and is halved at 80K, 120K and 160K iterations. The  training time is about 2 days on 2 Tesla V100 GPUs.

\subsection{Camera Pose Estimation}
\label{sec:pose}

We evaluate our pose estimation results (i.e., the predicted trajectory) qualitative and quantitatively. 
Specifically, we use 30-frame video clips from the RealEstate10K testing set, which consists of videos unseen during training. 
For each video clip, we estimate the relative pose between every two video frames and chain them together to get the full trajectory. 
Because this estimated transformation is in the coordinate space of deep voxels, we apply Umeyama alignment~\cite{umeyama1991least} to align the predicted camera trajectory with an SfM trajectory provided in the dataset.

We evaluate the Absolute Trajectory Error (ATE) on the RealEstate10K dataset and compare our performance with the state-of-the-art self-supervised viewpoint estimation method SSV~\cite{mustikovela2020self} and a structurally similar method SfMLearner~\cite{zhou2017unsupervised}. Additionally, we also compare with P$^2$-Net~\cite{IndoorSfMLearner} (a.k.a Indoor SfMlearner), an improved version of \cite{zhou2017unsupervised} that is optimized for indoor environment. We use 1000 30-frame (2.5-sec) video sequences and measure the difference between the trajectory estimated by each method and the trajectory obtained from SfM. Results are shown in Table~\ref{tab:ate}. Our result drastically reduces the error rate of the learning-based baseline method~\cite{zhou2017unsupervised} with about 69\% less in mean error and 72\% less in maximum error, suggesting that our approach learns much better viewpoint representations. Comparing to the Structure from Motion pipeline COLMAP~\cite{schoenberger2016sfm}, our method can obtain higher accuracy under the 30-frame testing setup. We also looked at the subset of clips on which COLMAP fails ($12.0\%$ of all clips): for these, our method achieves even better results ($\text{mean error}=0.004$, $\text{max error}=0.009$). Inspecting the failed videos, most have either very small motion, or pure rotations. These cases are hard/impossible for SfM but are potentially easy to learn. Finally, note COLMAP takes much longer to process a video compared to our method (71.53 secs versus our 0.004 secs). 

A further application of our camera trajectory is \textit{camera stabilization}. We show that we can warp future frames back into the viewpoint of the first frame by using the estimated relative pose between these two frames. Please see our website for details.

\begin{table}[t]
\centering
\tablestyle{6pt}{1}
\begin{tabular}{l|x{24}x{24}x{32}|c}
\multicolumn{1}{c|}{Method}  &  Mean$\downarrow$ & RMSE$\downarrow$  & Max err.$\downarrow$ & Density$\uparrow$  \\
\shline
SSV~\cite{mustikovela2020self} &  0.142 & 0.175 & 0.365 & $100.0\%$\\
P$^2$-Net~\cite{IndoorSfMLearner} & 0.059  & 0.068 & 0.1475 & $100.0\%$ \\
SFMLearner~\cite{zhou2017unsupervised} & 0.048  & 0.055 & 0.1105 & $100.0\%$ \\
COLMAP~\cite{schoenberger2016sfm} & 0.024 & 0.030 & 0.0765 & $88.0\%$ \\
\hline
Ours  & \textbf{0.017} & \textbf{0.019}  & \textbf{0.0410} & $100.0\%$   \\
\end{tabular}
\vspace{-0.1in}
\caption{Absolute Trajectory Error (ATE) on RealEstate10K~\cite{zhou2018stereo} dataset. We evaluate on 1000 30-frame video clips and take an average across all clips. }
\label{tab:ate}
\vspace{-0.2in}
\end{table}

\begin{figure*}[t]
    \centering
    \includegraphics[width=\textwidth]{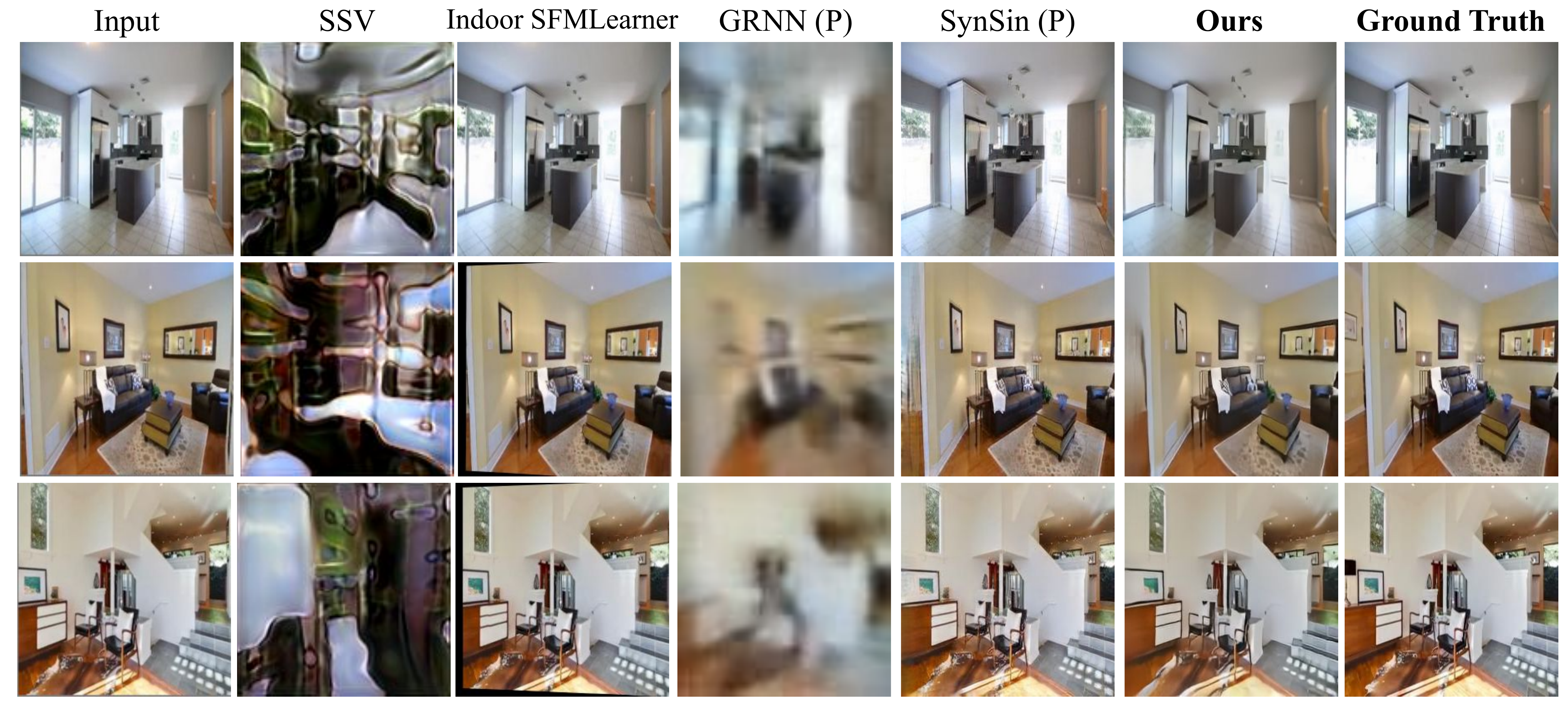}
    \vspace{-0.3in}
    \caption{\textbf{Novel View Synthesis (our method vs. previous methods):} Other methods show systematic errors either in rendering or pose estimation. Similar to our work, SSV~\cite{mustikovela2020self} make use of the least supervision signals. However, the view synthesis results are quite unsatisfactory. Indoor SFMLearner~\cite{IndoorSfMLearner} (SFMLearner optimized for indoor scenes) warps the image with predicted depth and pose. However, this warping operation could also cause large blank areas where no corresponding original pixels could be found.  GRNN~\cite{Tung_2019_CVPR} shares the most similar representation with ours, but it fails to generate clear images for reasons including their model could only handle 2 dof of camera transformation. Synsin~\cite{wiles2020synsin} shows the most competitive performance as their model trained with much stronger supervision. See Fig.~\ref{fig:details} for a detailed comparison. (P) denotes model trained with camera pose.}
    \vspace{-0.15in}
    \label{fig:qualitative_other}
\end{figure*}

\subsection{Novel View Synthesis}
\label{sec:viewsynthesis}

Creating a 3D walk-though from a single still image has been a classic computer graphics task~\cite{hoiem2005automatic,saxena2008make3d}, more recently known as single image Novel View Synthesis~\cite{tatarchenko2016multi,zhou2016view,  zhou2017unsupervised,wiles2020synsin,single_view_mpi}.
Given an image and a desired viewpoint, the aim is to synthesize the same scene from that new viewpoint.  We should that our video autoencoder can be effectively utilized for this task. 
We report results on two public datasets: RealEstate10K~\cite{zhou2018stereo} and Matterport3D~\cite{Matterport3D}. Additionally, we benchmark the generalization ability of our approach and the baselines on an additional dataset: Replica dataset~\cite{replica19arxiv} . We use the Replica datasets to test the out-of-domain accuracy of our model because they are not used during training.

\textbf{Metrics.} We compare against other methods using PSNR (Peak Signal-to-Noise Ratio), SSIM (Structural Similarity), and LPIPS (Perceptual Similarity). PSNR measures pixel-wise differences between two images, and SSIM measures luminance, contrast, and structure changes and aims to better reflect the human perceptual quality. A higher number indicates better results. LPIPS measures the distance in deep feature space, and is shown to be a good indicator of perceptual similarity. Lower values are better.

\begin{figure}[t]
    \centering
    \includegraphics[width=\linewidth]{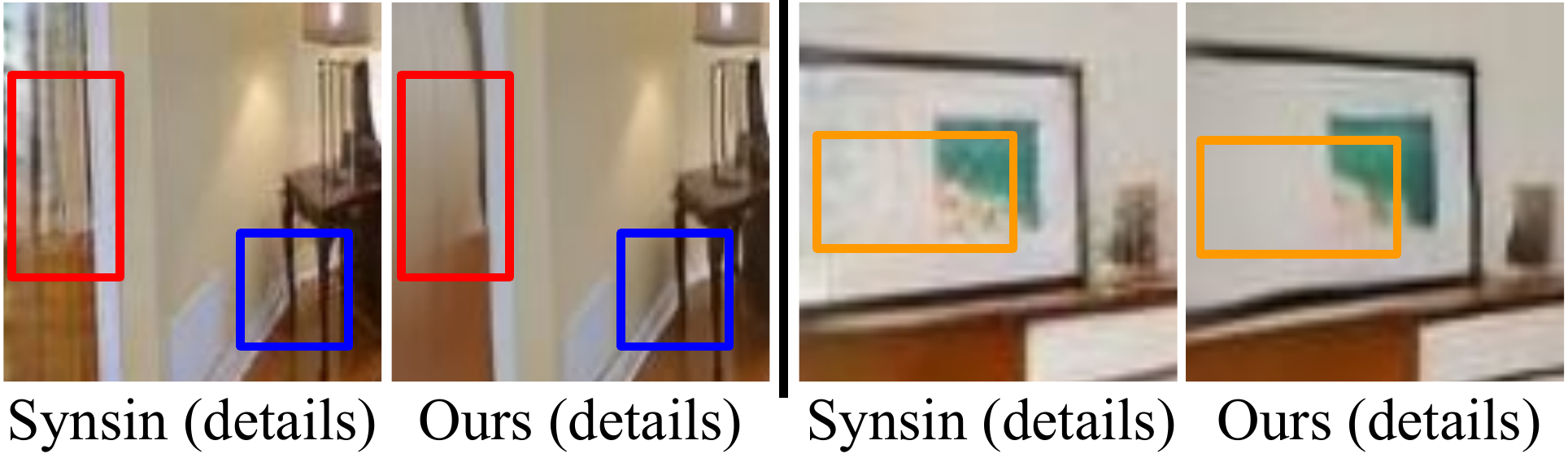}
    \vspace{-0.25in}
    \caption{\textbf{Extrapolating into unseen areas (details)}: Both Synsin and our model exhibit artifacts when extrapolating into unseen areas, but our model is able to produce more smoothed results (as shown in three colorful rectangles).}
    \vspace{-0.2in}
    \label{fig:details}
\end{figure}

\vspace{-0.05in}

\subsubsection{Novel View Synthesis on RealEstate10K}
\vspace{-0.05in}

The RealEstate10K dataset consists of footages of real estates (both indoor and outdoor) and is mostly static. We use 10000 videos for training and 5000 videos for testing. In Table~\ref{tab:RE10K}, we compare our method with previous approaches on the real dataset RealEstate10K. As seen in the table, we compare favorably to most single-image view synthesis algorithms, even though our method does not train on camera pose ground-truths while other methods do. Our method is able to achieve better results in both PSNR and SSIM than Synsin~\cite{wiles2020synsin}, which is a recent approach using point clouds as the intermediate representation and ground-truth cameras during training. This indicates learning the disentangled deep voxel and camera pose jointly can lead to better 3D representations. 
Our method also easily outperforms methods trained without using camera poses. SSV~\cite{mustikovela2020self}, which similarly uses no camera information during training, fails to generate meaningful content. SfMLearner~\cite{zhou2017unsupervised} and the subsequent P$^2$-Net~\cite{IndoorSfMLearner} (a.k.a. Indoor SfMLearner) are structurally similar to ours. We warp the image input with the predicted depth and pose to test their view synthesis results. However, these methods are not optimized for view synthesis and fail to achieve the image quality as ours. 

Figure~\ref{fig:qualitative_other} shows the qualitative comparisons. Given a \emph{single image} and a specified motion trajectory as inputs, our method is able to generate photorealistic results with correct motion. While Synsin~\cite{wiles2020synsin} achieves competitive results, it requires true camera poses for training. Our model also shows reasonable extrapolation into unseen areas. Both row 2 and 3 involve extrapolating into unknown areas to the left. Fig.~\ref{fig:details} shows detailed comparison in unseen areas between our model and \cite{wiles2020synsin}. Our model produces more smoothed results and \cite{wiles2020synsin} shows stronger artifacts. 

\begin{table}[t]
\centering
\tablestyle{5pt}{1.0}

\begin{tabular}{l|cc|ccc}
\multicolumn{1}{c|}{Method}  & Cam$_{\text{in}}$  & Cam$_{\text{ex}}$ & PSNR$\uparrow$ & SSIM$\uparrow$ & LPIPS$\downarrow$ \\
\shline
\multicolumn{6}{l}{\emph{methods without any camera supervision:}} \\
\hline
SSV\cite{mustikovela2020self}  & × & × & 7.95  & 0.19 & 4.12  \\
\textbf{Ours}  & × & ×  & \textbf{23.21} & \textbf{0.73}  & \textbf{1.54} \\
\hline
\multicolumn{6}{l}{\emph{methods trained with camera intrinsics:}} \\
\hline
SfMLearner\cite{zhou2017unsupervised}& \checkmark & ×  & 15.82 & 0.46 & 2.39\\
MonoDepth2\cite{godard2019digging} & \checkmark & × & 17.15  & 0.55 & 2.08 \\
P$^2$-Net\cite{IndoorSfMLearner} & \checkmark & × & 17.77 & 0.56 & 1.96\\
\hline
\multicolumn{6}{l}{\emph{methods trained with camera intrinsics and extrinsics:}} \\
\hline
Dosovitsky et al~\cite{dosovitskiy2015learning} & \checkmark& \checkmark & 11.35 & 0.33 & 3.95\\
GQN~\cite{eslami2018neural} & \checkmark & \checkmark& 16.94 & 0.56 & 3.33 \\
Appearance Flow~\cite{zhou2016view} & \checkmark & \checkmark& 17.05 & 0.56 & 2.19\\
GRNN~\cite{Tung_2019_CVPR} & \checkmark & \checkmark& 19.13 & 0.63 & 2.83 \\
3DPaper~\cite{wiles2020synsin} & \checkmark & \checkmark& 21.88 & 0.66 & 1.52\\
SynSin (w/ voxel)~\cite{wiles2020synsin} & \checkmark& \checkmark & 21.88 & 0.71 & 1.30\\
SynSin~\cite{wiles2020synsin} & \checkmark& \checkmark & 22.31 & 0.74 & \textbf{1.18}\\
Single-view MPIs~\cite{single_view_mpi} & \checkmark& \checkmark & 23.70 & 0.80 & - \\
StereoMag$^\dagger$~\cite{zhou2018stereo} & \checkmark& \checkmark & \textbf{25.34} & \textbf{0.82} &  1.19\\
\end{tabular}
\vspace{-0.1in}
\caption{Novel view synthesis task with RealEstate10K~\cite{zhou2018stereo}. We follow the standard metrics of PSNR, SSIM and LPIPS~\cite{zhang2018unreasonable}. For PSNR and SSIM, higher numbers are better. For LPIPS, lower numbers are better. We use implementation of \cite{wiles2020synsin} to compute LPIPS. $^\dagger$ StereoMag makes use of 2 images as input.}
\label{tab:RE10K}
\vspace{-0.05in}
\end{table}

\vspace{-0.15in}
\subsubsection{Novel View Synthesis on Matterport3D}
\vspace{-0.05in}

\begin{figure}[t]
    \centering
    \includegraphics[width=\linewidth]{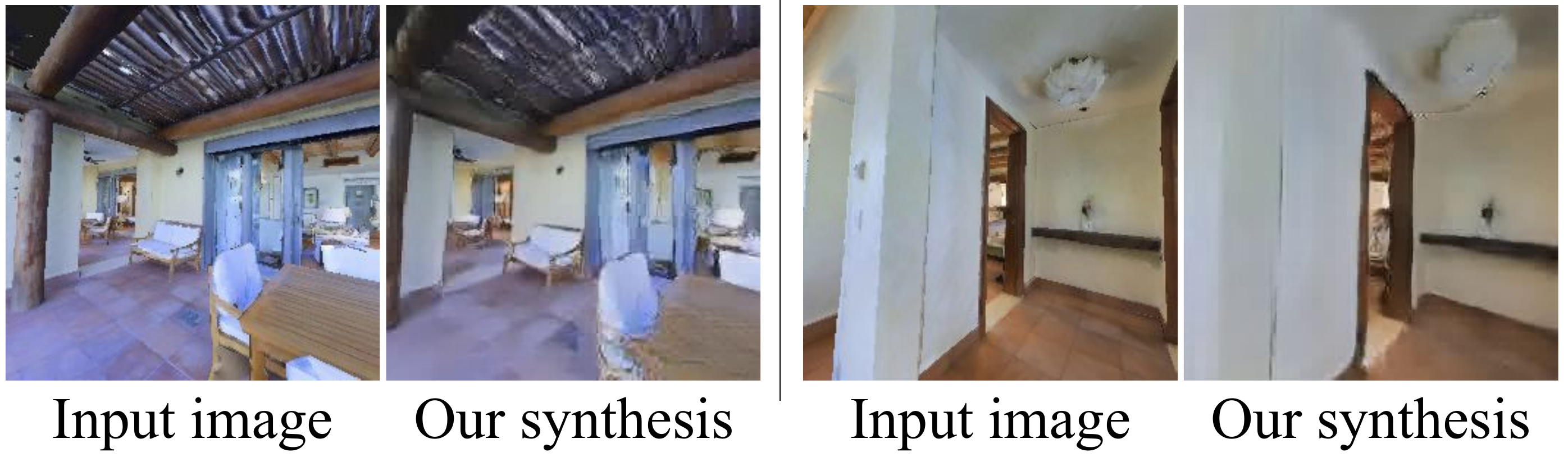}
    \vspace{-0.25in}
    \caption{Our model can also be applied on  Matterport3D~\cite{Matterport3D}, using images rendered from 3D models in the dataset.}
    \vspace{-0.15in}
    \label{fig:mp3d}
\end{figure}

\begin{table}[t]
\centering
\tablestyle{7.5pt}{1.0}

\begin{tabular}{l|c|cccc}
\multicolumn{1}{c|}{Method}  & Pose  & PSNR$\uparrow$& SSIM$\uparrow$ & LPIPS$\downarrow$\\
\shline
\multicolumn{5}{l}{\emph{methods without any camera supervision:}} \\
\hline
\textbf{Ours}  & × & 20.58 & 0.64 & 2.44 \\
\hline
\multicolumn{5}{l}{\emph{methods trained with camera intrinsics and extrinsics:}} \\
\hline
Dosovitsky et al~\cite{dosovitskiy2015learning} & \checkmark & 14.79 & 0.57 & 3.73\\
Appearance Flow~\cite{zhou2016view} & \checkmark & 15.87 & 0.53 & 2.99\\
Synsin (w/ voxel)~\cite{wiles2020synsin} & \checkmark & 20.62 & 0.70 & 1.97\\
Synsin~\cite{wiles2020synsin} & \checkmark & 20.91 & 0.72 & 1.68\\
\end{tabular}
\vspace{-0.05in}
\caption{Novel view synthesis with Matterport3D~\cite{Matterport3D}. We follow the standard metrics of PSNR and SSIM. Higher values are better.}
\label{tab:mp3d}
\end{table}

We evaluate the Video Autoencoder on the Matterport3D dataset~\cite{Matterport3D}. The Matterport3D dataset is a collection of 3D models of scanned and reconstructed properties. It consists of 61 training scenes and 18 testing scenes. We use a navigation agent in the Habitat simulator~\cite{habitat19iccv} to render around 100 episodes per scene as videos. These videos show an agent navigating from one point in the scene to another point. A total of around 6000 videos is generated for training and 800 videos for testing.  For experiments, we rendered an additional 60K image pairs related by a random rotation and translation for fine-tuning the model. In this way, the training data includes rotations in all three axes.

Table~\ref{tab:mp3d} shows the numerical results for Video Autoencoder on the Matterport3D. Without training on any camera information, our model is able to perform comparably to methods supervised with pose across all three metrics.  Figure~\ref{fig:mp3d} visualizes the view synthesis results of our method using data from the Matterport3D dataset. Although the visual appearance rendered from 3D models is usually flawed due to incomplete point clouds, occlusions, and other possible artifacts, our model still produces satisfactory results.

\vspace{-0.1in}
\subsubsection{Generalization Ability of Novel View Synthesis}
\vspace{-0.05in}
We benchmark Video Autoencoder on an out-of-domain dataset, Replica~\cite{replica19arxiv}, a set of 3D reconstructions of indoor spaces, to evaluate the generalization ability of our model, without any further finetuning. We rendered 200 image pairs on each of the 5 episodes with Habitat simulator~\cite{habitat19iccv} as the test set. Table~\ref{tab:replica} shows numerical results compared to other methods. All methods use Matterport3D as the training dataset and only test on Replica without any finetuning on it. We observe the same trend as in other datasets, with our method perform comparably to existing methods that require camera supervision. This indicates that our method is able to consistently outperform the baseline methods on generalization to out-of-domain data.

\begin{table}[t]
\centering
\tablestyle{7.5pt}{1.0}
\begin{tabular}{l|c|cccc}
\multicolumn{1}{c|}{Method}  & Pose  & PSNR$\uparrow$& SSIM$\uparrow$ & LPIPS$\downarrow$\\
\shline
\multicolumn{5}{l}{\emph{methods without any camera supervision:}} \\
\hline
\textbf{Ours}  & × & 21.72 & 0.77 & 2.21  \\
\hline
\multicolumn{5}{l}{\emph{methods trained with camera intrinsics and extrinsics:}} \\
\hline
Dosovitsky et al~\cite{dosovitskiy2015learning} & \checkmark & 14.36 & 0.68 & 3.36 \\
Appearance Flow~\cite{zhou2016view} & \checkmark & 17.42 & 0.66 & 2.29\\
Synsin (w/ voxel)~\cite{wiles2020synsin} & \checkmark & 19.77 & 0.75 & 2.24 \\
Synsin~\cite{wiles2020synsin} & \checkmark & 21.94 & 0.81 & 1.55 \\
\end{tabular}
\vspace{-0.1in}
\caption{Novel view synthesis task with Replica dataset~\cite{replica19arxiv}.}
\label{tab:replica}
\vspace{-0.2in}
\end{table}

\begin{table*}[t]\centering\vspace{-3mm}
\subfloat[\textbf{3D deep voxel spatial resolution}: Our final model with a resolution of $64\times 64$ offers the best performance. \label{tab:ablation:resolution}]{
\tablestyle{5pt}{1}
\begin{tabular}{c|x{22}x{22}x{22}}
\multicolumn{1}{c|}{Resolution}  & SSIM$\uparrow$ & PSNR$\uparrow$ & LPIPS$\downarrow$ \\
\shline
128\x 128 & 0.70 & 22.39 & 1.63 \\
64\x 64 & \textbf{0.73} & \textbf{23.21}  & \textbf{1.54} \\
32\x 32 &  0.61 & 20.23 & 2.77 \\
\end{tabular}}\hspace{3mm}
\subfloat[\textbf{3D deep voxel depth resolution}: A depth dimension of 32 shows superior results compared to other variants. \label{tab:ablation:depth}]{
\tablestyle{5pt}{1}
\begin{tabular}{c|x{22}x{22}x{22}}
\multicolumn{1}{c|}{Depth} & SSIM$\uparrow$ & PSNR$\uparrow$ & LPIPS$\downarrow$ \\
\shline
D=64 & 0.71 & 22.05 & 1.61 \\
D=32 & \textbf{0.73} & \textbf{23.21}  & \textbf{1.54} \\
D=16 & 0.70 & 21.92 & 1.86  \\
\end{tabular}}\hspace{3mm}
\subfloat[\textbf{Video clip for training - number of frame}: Training with a clip of 6 frames can offer better overall performance . \label{tab:ablation:nframe}]{
\tablestyle{5pt}{1}
\begin{tabular}{l|x{22}x{22}x{22}}
\multicolumn{1}{c|}{\# Frame}  & SSIM$\uparrow$ & PSNR$\uparrow$ & LPIPS$\downarrow$\\
\shline
3 frames & 0.72 & 22.52 & \textbf{1.52}   \\
6 frames & \textbf{0.73} & \textbf{23.21}  & {1.54} \\
10 frames & 0.69 & 21.54 & 1.73  \\
\end{tabular}}\hspace{3mm}
\vspace{-0.1in}
\caption{\textbf{Ablations} on RealEstate10K view synthesis results. We show SSIM, PSNR and LPIPS performance on testing set.}
\vspace{-0.1in}
\label{tab:ablations}
\end{table*}
\begin{figure*}[t]
    \centering
    \includegraphics[width=.9\textwidth]{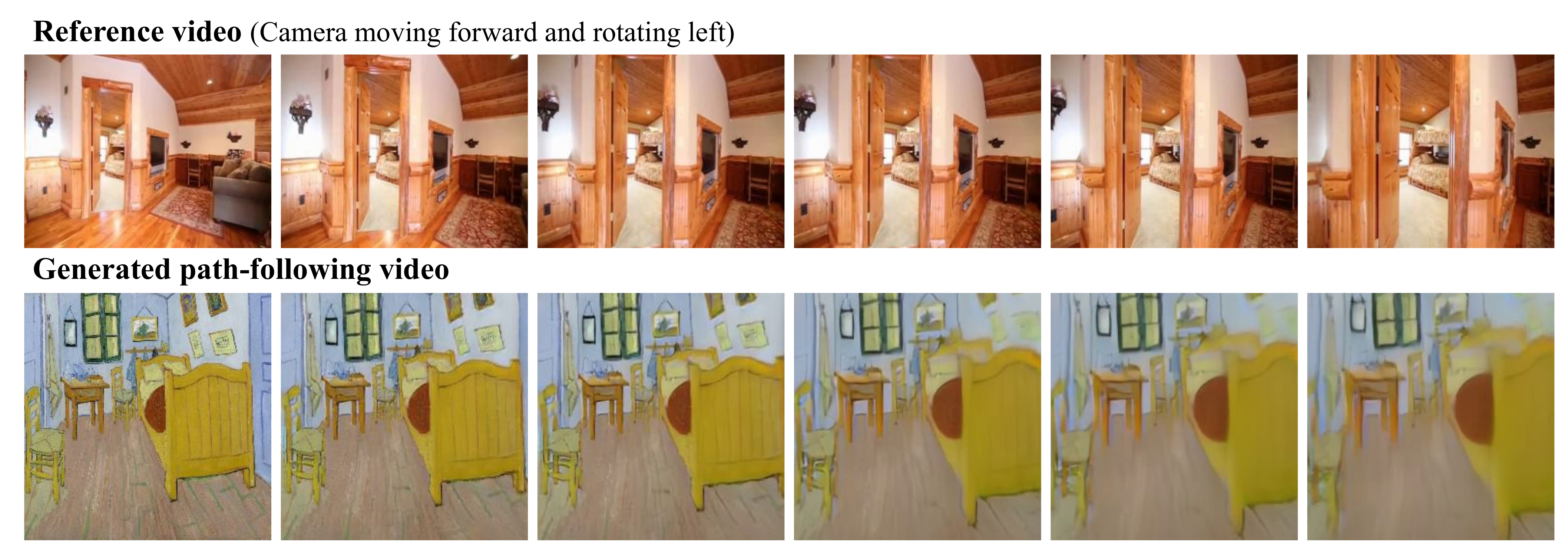}
    \vspace{-0.1in}
    \caption{\textbf{Following trajectories of other videos:} by using trajectories of other videos, we can even animate oil paintings.}
    \vspace{-0.1in}
    \label{fig:vidfollow}
\end{figure*}

\vspace{-0.05in}
\subsubsection{Ablation Studies}
\vspace{-0.05in}
To quantitatively evaluate the impact of various components of the Video Autoencoder, we conduct a set of ablation studies using variants of our model in Table~\ref{tab:ablations}. The best model with 6 input frames, $64 \times 64$ spatial resolution and $32$ depth resolution in the 3D voxel corresponds to our model reported in previous sections. 

\textbf{Resolution of 3D voxels: } We compare our default voxel resolution with variants that modify the encoder's output spatial resolution and depth.  As shown in Table~\ref{tab:ablation:resolution}, the errors increase drastically when the spatial resolution is changed. 
Comparing to models with depth changed (Table~\ref{tab:ablation:depth}), although the other variants are capable of attaining reasonable performance, our default model achieves the best performance.

\textbf{Video clip for training:} We modify the training video clip by changing the clip length. As seen in Table \ref{tab:ablation:nframe}, the performance improved when the clip length is increase. We conjecture that a clip too short could provide an appropriate scale of motion, which is crucial for training. However, if we further expand the clip length to 10 frames, the performance drops by about $5\%$. We hypothesize that predicting the last frame from the first frame becomes too difficult under such a setting, which is undesirable for training. This suggests that 6-frame clips are more effective training data.

\subsection{Video Following}
\label{sec:videofollowing}
Finally, we evaluate Video Autoencoder on the task of animating a single image with the motion trajectories from different videos. Specifically, we obtain a 3D deep voxels representation from our desired image and trajectory from a different video. We then combine the trajectory and the 3D structure for the decoder to render a new video.

Figure~\ref{fig:vidfollow} visualizes a video predicted from an out-of-domain image. Training only once on the RealEstate10K dataset, our model can adapt to a diverse set of images. 
The shown frames are generated from the painting \textit{Bedroom in Arles}. Although the painting has a texture that is quite different from the training dataset, our method still models it reasonably well. Adapting from the reference video, the generated sequence shows a trajectory as if we are walking into Vincent van Gogh's bedroom in Arles.

\vspace{-0.05in}
\section{Conclusion}
\vspace{-0.05in}
We present Video Autoencoder that encodes videos into disentangled representations of 3D structure and camera pose. The model is trained with only raw videos without using any explicit 3D supervision or camera pose. We show that our representation enables tasks such as camera pose estimation, novel view synthesis and video generation by motion following. Our model demonstrates superior generalization ability on all tasks and achieves state-of-the-art results on self-supervised camera pose estimation. Our model also achieves on par results on novel view synthesis comapred to approaches using ground-truth camera in training.

{\footnotesize \textbf{Acknowledgements.}~This work was supported, in part, by grants from DARPA MCS and LwLL, NSF 1730158 CI-New: Cognitive Hardware and Software Ecosystem Community Infrastructure, NSF ACI-1541349 CC*DNI Pacific Research Platform, and gifts from Qualcomm, TuSimple and Picsart. We thank Taesung Park and Bill Peebles for valuable comments.}

{\small
\bibliographystyle{ieee_fullname}
\bibliography{shortstrings,egbib,zihang}

\begin{thebibliography}{10}\itemsep=-1pt

\bibitem{agarwal2009building}
Sameer Agarwal, Noah Snavely, Ian Simon, Steven~M Seitz, and Richard Szeliski.
\newblock Building rome in a day.
\newblock In {\em Proc. ICCV}, 2009.

\bibitem{agrawal2015learning}
Pulkit Agrawal, Joao Carreira, and Jitendra Malik.
\newblock Learning to see by moving.
\newblock In {\em Proceedings of the IEEE international conference on computer
  vision}, pages 37--45, 2015.

\bibitem{anokhin2020high}
Ivan Anokhin, Pavel Solovev, Denis Korzhenkov, Alexey Kharlamov, Taras
  Khakhulin, Aleksei Silvestrov, Sergey Nikolenko, Victor Lempitsky, and Gleb
  Sterkin.
\newblock High-resolution daytime translation without domain labels.
\newblock In {\em Proceedings of the IEEE/CVF Conference on Computer Vision and
  Pattern Recognition}, pages 7488--7497, 2020.

\bibitem{bansal20204d}
Aayush Bansal, Minh Vo, Yaser Sheikh, Deva Ramanan, and Srinivasa Narasimhan.
\newblock 4d visualization of dynamic events from unconstrained multi-view
  videos.
\newblock In {\em Proc. CVPR}, 2020.

\bibitem{barrow1978recovering}
Harry~G. Barrow and J.M. Tenenbaum.
\newblock Recovering intrinsic scene characteristics.
\newblock {\em Comput. Vis. Syst}, 2(3-26):2, 1978.

\bibitem{Matterport3D}
Angel Chang, Angela Dai, Thomas Funkhouser, Maciej Halber, Matthias Niessner,
  Manolis Savva, Shuran Song, Andy Zeng, and Yinda Zhang.
\newblock Matterport3d: Learning from rgb-d data in indoor environments.
\newblock {\em 3}.

\bibitem{shapenet2015}
Angel~X. Chang, Thomas Funkhouser, Leonidas Guibas, Pat Hanrahan, Qixing Huang,
  Zimo Li, Silvio Savarese, Manolis Savva, Shuran Song, Hao Su, Jianxiong Xiao,
  Li Yi, and Fisher Yu.
\newblock {ShapeNet: An Information-Rich 3D Model Repository}.
\newblock Technical Report arXiv:1512.03012 [cs.GR], Stanford University ---
  Princeton University --- Toyota Technological Institute at Chicago, 2015.

\bibitem{chen2016infogan}
Xi Chen, Yan Duan, Rein Houthooft, John Schulman, Ilya Sutskever, and Pieter
  Abbeel.
\newblock Infogan: Interpretable representation learning by information
  maximizing generative adversarial nets.
\newblock In {\em Advances in neural information processing systems}, pages
  2172--2180, 2016.

\bibitem{denton2017unsupervised}
Emily Denton and Vighnesh Birodkar.
\newblock Unsupervised learning of disentangled representations from video.
\newblock In {\em Proceedings of the 31st International Conference on Neural
  Information Processing Systems}, pages 4417--4426, 2017.

\bibitem{dosovitskiy2016inverting}
Alexey Dosovitskiy and Thomas Brox.
\newblock Inverting visual representations with convolutional networks.
\newblock In {\em Proceedings of the IEEE conference on computer vision and
  pattern recognition}, pages 4829--4837, 2016.

\bibitem{dosovitskiy2015learning}
Alexey Dosovitskiy, Jost Tobias~Springenberg, and Thomas Brox.
\newblock Learning to generate chairs with convolutional neural networks.
\newblock In {\em Proc. CVPR}, 2015.

\bibitem{eslami2018neural}
SM~Ali Eslami, Danilo~Jimenez Rezende, Frederic Besse, Fabio Viola, Ari~S
  Morcos, Marta Garnelo, Avraham Ruderman, Andrei~A Rusu, Ivo Danihelka, Karol
  Gregor, et~al.
\newblock Neural scene representation and rendering.
\newblock {\em Science}, 360(6394):1204--1210, 2018.

\bibitem{Tung_2019_CVPR}
Hsiao-Yu Fish~Tung, Ricson Cheng, and Katerina Fragkiadaki.
\newblock Learning spatial common sense with geometry-aware recurrent networks.
\newblock In {\em cvpr}.

\bibitem{flynn2016deepstereo}
J. {Flynn}, I. {Neulander}, J. {Philbin}, and N. {Snavely}.
\newblock Deep stereo: Learning to predict new views from the world's imagery.
\newblock In {\em 2016 IEEE Conference on Computer Vision and Pattern
  Recognition (CVPR)}, pages 5515--5524, 2016.

\bibitem{godard2019digging}
Cl{\'e}ment Godard, Oisin Mac~Aodha, Michael Firman, and Gabriel~J Brostow.
\newblock Digging into self-supervised monocular depth estimation.
\newblock In {\em Proceedings of the IEEE/CVF International Conference on
  Computer Vision}, pages 3828--3838, 2019.

\bibitem{gulrajani2017improved}
Ishaan Gulrajani, Faruk Ahmed, Martin Arjovsky, Vincent Dumoulin, and Aaron~C
  Courville.
\newblock Improved training of wasserstein gans.
\newblock In {\em NerIPS}, 2017.

\bibitem{Han19}
Tengda Han, Weidi Xie, and Andrew Zisserman.
\newblock Video representation learning by dense predictive coding.
\newblock In {\em Proc. ICCV}, 2019.

\bibitem{harley2020tracking}
Adam~W Harley, Shrinidhi~Kowshika Lakshmikanth, Paul Schydlo, and Katerina
  Fragkiadaki.
\newblock Tracking emerges by looking around static scenes, with neural 3d
  mapping.
\newblock In {\em eccv}, 2020.

\bibitem{harley_viewcontrast}
Adam~W. Harley, Fangyu Li, Shrinidhi~K. Lakshmikanth, Xian Zhou, Hsiao-Yu~Fish
  Tung, and Katerina Fragkiadaki.
\newblock Learning from unlabelled videos using contrastive predictive neural
  3d mapping.
\newblock In {\em ICLR}, 2019.

\bibitem{harsh2018disentangling}
Ananya Harsh~Jha, Saket Anand, Maneesh Singh, and VSR Veeravasarapu.
\newblock Disentangling factors of variation with cycle-consistent variational
  auto-encoders.
\newblock In {\em Proceedings of the European Conference on Computer Vision
  (ECCV)}, pages 805--820, 2018.

\bibitem{hartley2003multiple}
Richard Hartley and Andrew Zisserman.
\newblock {\em Multiple view geometry in computer vision}.
\newblock Cambridge university press, 2003.

\bibitem{he2016deep}
Kaiming He, Xiangyu Zhang, Shaoqing Ren, and Jian Sun.
\newblock Deep residual learning for image recognition.
\newblock In {\em Proc. CVPR}, 2016.

\bibitem{heljakka2020deep}
Ari Heljakka, Yuxin Hou, Juho Kannala, and Arno Solin.
\newblock Deep automodulators.
\newblock {\em Advances in Neural Information Processing Systems}, 33, 2020.

\bibitem{higgins2016beta}
Irina Higgins, Loic Matthey, Arka Pal, Christopher Burgess, Xavier Glorot,
  Matthew Botvinick, Shakir Mohamed, and Alexander Lerchner.
\newblock beta-vae: Learning basic visual concepts with a constrained
  variational framework.
\newblock 2016.

\bibitem{hoiem2005automatic}
Derek Hoiem, Alexei~A Efros, and Martial Hebert.
\newblock Automatic photo pop-up.
\newblock In {\em ACM SIGGRAPH 2005 Papers}. 2005.

\bibitem{hsieh2018learning}
Jun-Ting Hsieh, Bingbin Liu, De-An Huang, Li~F Fei-Fei, and Juan~Carlos
  Niebles.
\newblock Learning to decompose and disentangle representations for video
  prediction.
\newblock In {\em Advances in Neural Information Processing Systems}, pages
  517--526, 2018.

\bibitem{hu2021worldsheet}
Ronghang Hu, Nikhila Ravi, Alexander~C. Berg, and Deepak Pathak.
\newblock Worldsheet: Wrapping the world in a 3d sheet for view synthesis from
  a single image.
\newblock In {\em Proceedings of the IEEE International Conference on Computer
  Vision (ICCV)}, 2021.

\bibitem{huang2018munit}
Xun Huang, Ming-Yu Liu, Serge Belongie, and Jan Kautz.
\newblock Multimodal unsupervised image-to-image translation.
\newblock In {\em ECCV}, 2018.

\bibitem{Isola15}
Phillip Isola, Daniel Zoran, Dilip Krishnan, and Edward~H. Adelson.
\newblock Learning visual groups from co-occurrences in space and time.
\newblock In {\em Proc. ICLR}, 2015.

\bibitem{jakab2018unsupervised}
Tomas Jakab, Ankush Gupta, Hakan Bilen, and Andrea Vedaldi.
\newblock Unsupervised learning of object landmarks through conditional image
  generation.
\newblock In {\em Advances in neural information processing systems}, pages
  4016--4027, 2018.

\bibitem{Jayaraman15}
Dinesh Jayaraman and Kristen Grauman.
\newblock Learning image representations tied to ego-motion.
\newblock In {\em Proc. ICCV}, 2015.

\bibitem{Jayaraman16}
Dinesh Jayaraman and Kristen Grauman.
\newblock Slow and steady feature analysis: higher order temporal coherence in
  video.
\newblock In {\em Proc. CVPR}, 2016.

\bibitem{jimenez2016unsupervised}
Danilo Jimenez~Rezende, SM Eslami, Shakir Mohamed, Peter Battaglia, Max
  Jaderberg, and Nicolas Heess.
\newblock Unsupervised learning of 3d structure from images.
\newblock {\em Advances in neural information processing systems},
  29:4996--5004, 2016.

\bibitem{johnson2016perceptual}
Justin Johnson, Alexandre Alahi, and Li Fei-Fei.
\newblock Perceptual losses for real-time style transfer and super-resolution.
\newblock In {\em European conference on computer vision}, pages 694--711.
  Springer, 2016.

\bibitem{kanazawa2018learning}
Angjoo Kanazawa, Shubham Tulsiani, Alexei~A Efros, and Jitendra Malik.
\newblock Learning category-specific mesh reconstruction from image
  collections.
\newblock In {\em Proc. ECCV}, 2018.

\bibitem{karras2019style}
Tero Karras, Samuli Laine, and Timo Aila.
\newblock A style-based generator architecture for generative adversarial
  networks.
\newblock In {\em Proceedings of the IEEE conference on computer vision and
  pattern recognition}, pages 4401--4410, 2019.

\bibitem{kim2018disentangling}
Hyunjik Kim and Andriy Mnih.
\newblock Disentangling by factorising.
\newblock In {\em ICML}, 2018.

\bibitem{kingma2014adam}
Diederik~P Kingma and Jimmy Ba.
\newblock Adam: A method for stochastic optimization.
\newblock {\em arXiv preprint arXiv:1412.6980}, 2014.

\bibitem{kingma2013auto}
Diederik~P Kingma and Max Welling.
\newblock Auto-encoding variational bayes.
\newblock {\em arXiv preprint arXiv:1312.6114}, 2013.

\bibitem{kulkarni2015deep}
Tejas~D Kulkarni, William~F Whitney, Pushmeet Kohli, and Josh Tenenbaum.
\newblock Deep convolutional inverse graphics network.
\newblock In {\em Advances in neural information processing systems}, pages
  2539--2547, 2015.

\bibitem{lee2020drit}
Hsin-Ying Lee, Hung-Yu Tseng, Qi Mao, Jia-Bin Huang, Yu-Ding Lu, Maneesh Singh,
  and Ming-Hsuan Yang.
\newblock Drit++: Diverse image-to-image translation via disentangled
  representations.
\newblock {\em International Journal of Computer Vision}, pages 1--16, 2020.

\bibitem{liu2020factorize}
Andrew Liu, Shiry Ginosar, Tinghui Zhou, Alexei~A. Efros, and Noah Snavely.
\newblock Learning to factorize and relight a city.
\newblock In {\em ECCV}, 2020.

\bibitem{mildenhall2020nerf}
Ben Mildenhall, Pratul~P Srinivasan, Matthew Tancik, Jonathan~T Barron, Ravi
  Ramamoorthi, and Ren Ng.
\newblock Nerf: Representing scenes as neural radiance fields for view
  synthesis.
\newblock In {\em Proc. ECCV}, 2020.

\bibitem{minderer2019unsupervised}
Matthias Minderer, Chen Sun, Ruben Villegas, Forrester Cole, Kevin~P Murphy,
  and Honglak Lee.
\newblock Unsupervised learning of object structure and dynamics from videos.
\newblock In {\em Advances in Neural Information Processing Systems}, pages
  92--102, 2019.

\bibitem{Misra16}
Ishan Misra, C.~Lawrence Zitnick, and Martial Hebert.
\newblock Shuffle and learn: Unsupervised learning using temporal order
  verification.
\newblock In {\em Proc. ECCV}, 2016.

\bibitem{mustikovela2020self}
Siva~Karthik Mustikovela, Varun Jampani, Shalini~De Mello, Sifei Liu, Umar
  Iqbal, Carsten Rother, and Jan Kautz.
\newblock Self-supervised viewpoint learning from image collections.
\newblock In {\em Proceedings of the IEEE/CVF Conference on Computer Vision and
  Pattern Recognition}, pages 3971--3981, 2020.

\bibitem{nguyen2019hologan}
Thu Nguyen-Phuoc, Chuan Li, Lucas Theis, Christian Richardt, and Yong-Liang
  Yang.
\newblock Hologan: Unsupervised learning of 3d representations from natural
  images.
\newblock In {\em Proceedings of the IEEE International Conference on Computer
  Vision}, pages 7588--7597, 2019.

\bibitem{niklaus20193d}
Simon Niklaus, Long Mai, Jimei Yang, and Feng Liu.
\newblock 3d ken burns effect from a single image.
\newblock {\em ACM Trans. Gr.}, 38(6), 2019.

\bibitem{park2020swapping}
Taesung Park, Jun-Yan Zhu, Oliver Wang, Jingwan Lu, Eli Shechtman, Alexei
  Efros, and Richard Zhang.
\newblock Swapping autoencoder for deep image manipulation.
\newblock {\em Advances in Neural Information Processing Systems}, 33, 2020.

\bibitem{peebles2020hessian}
William Peebles, John Peebles, Jun-Yan Zhu, Alexei~A. Efros, and Antonio
  Torralba.
\newblock The hessian penalty: A weak prior for unsupervised disentanglement.
\newblock In {\em Proceedings of European Conference on Computer Vision
  (ECCV)}, 2020.

\bibitem{pidhorskyi2020adversarial}
Stanislav Pidhorskyi, Donald~A Adjeroh, and Gianfranco Doretto.
\newblock Adversarial latent autoencoders.
\newblock In {\em Proceedings of the IEEE/CVF Conference on Computer Vision and
  Pattern Recognition}, pages 14104--14113, 2020.

\bibitem{prabhudesai2021disentangling}
Mihir Prabhudesai, Shamit Lal, Darshan Patil, Hsiao-Yu Tung, Adam~W Harley, and
  Katerina Fragkiadaki.
\newblock Disentangling 3d prototypical networks for few-shot concept learning.
\newblock In {\em iclr}, 2021.

\bibitem{Rockwell2021}
Chris Rockwell, David~F. Fouhey, and Justin Johnson.
\newblock Pixelsynth: Generating a 3d-consistent experience from a single
  image.
\newblock In {\em ICCV}, 2021.

\bibitem{habitat19iccv}
Manolis Savva, Abhishek Kadian, Oleksandr Maksymets, Yili Zhao, Erik Wijmans,
  Bhavana Jain, Julian Straub, Jia Liu, Vladlen Koltun, Jitendra Malik, Devi
  Parikh, and Dhruv Batra.
\newblock Habitat: {A} {P}latform for {E}mbodied {AI} {R}esearch.
\newblock In {\em Proc. ICCV}, 2019.

\bibitem{saxena2008make3d}
Ashutosh Saxena, Min Sun, and Andrew~Y Ng.
\newblock Make3d: Learning 3d scene structure from a single still image.
\newblock {\em IEEE transactions on pattern analysis and machine intelligence},
  31(5):824--840, 2008.

\bibitem{schoenberger2016sfm}
Johannes~Lutz Sch\"{o}nberger and Jan-Michael Frahm.
\newblock Structure-from-motion revisited.
\newblock In {\em Proc. CVPR}, 2016.

\bibitem{shen2020interpreting}
Yujun Shen, Jinjin Gu, Xiaoou Tang, and Bolei Zhou.
\newblock Interpreting the latent space of gans for semantic face editing.
\newblock In {\em Proceedings of the IEEE/CVF Conference on Computer Vision and
  Pattern Recognition}, pages 9243--9252, 2020.

\bibitem{sitzmann2019deepvoxels}
Vincent Sitzmann, Justus Thies, Felix Heide, Matthias Nie{\ss}ner, Gordon
  Wetzstein, and Michael Zollhofer.
\newblock Deepvoxels: Learning persistent 3d feature embeddings.
\newblock In {\em Proc. CVPR}, 2019.

\bibitem{sitzmann2019scene}
Vincent Sitzmann, Michael Zollh{\"o}fer, and Gordon Wetzstein.
\newblock Scene representation networks: Continuous 3d-structure-aware neural
  scene representations.
\newblock In {\em NerIPS}, 2019.

\bibitem{snavely2008scene}
Keith~N Snavely.
\newblock {\em Scene reconstruction and visualization from internet photo
  collections}.
\newblock University of Washington USA, 2008.

\bibitem{replica19arxiv}
Julian Straub, Thomas Whelan, Lingni Ma, Yufan Chen, Erik Wijmans, Simon Green,
  Jakob~J. Engel, Raul Mur-Artal, Carl Ren, Shobhit Verma, Anton Clarkson,
  Mingfei Yan, Brian Budge, Yajie Yan, Xiaqing Pan, June Yon, Yuyang Zou,
  Kimberly Leon, Nigel Carter, Jesus Briales, Tyler Gillingham, Elias Mueggler,
  Luis Pesqueira, Manolis Savva, Dhruv Batra, Hauke~M. Strasdat, Renzo~De
  Nardi, Michael Goesele, Steven Lovegrove, and Richard Newcombe.
\newblock The {R}eplica dataset: A digital replica of indoor spaces.
\newblock {\em arXiv preprint arXiv:1906.05797}, 2019.

\bibitem{tatarchenko2016multi}
Maxim Tatarchenko, Alexey Dosovitskiy, and Thomas Brox.
\newblock Multi-view 3d models from single images with a convolutional network.
\newblock In {\em European Conference on Computer Vision}, pages 322--337.
  Springer, 2016.

\bibitem{tewari2020state}
Ayush Tewari, Ohad Fried, Justus Thies, Vincent Sitzmann, Stephen Lombardi,
  Kalyan Sunkavalli, Ricardo Martin-Brualla, Tomas Simon, Jason Saragih,
  Matthias Nie{\ss}ner, et~al.
\newblock State of the art on neural rendering.
\newblock In {\em Computer Graphics Forum}, volume~39, pages 701--727. Wiley
  Online Library, 2020.

\bibitem{tomasi1992shape}
Carlo Tomasi and Takeo Kanade.
\newblock Shape and motion from image streams under orthography: a
  factorization method.
\newblock {\em International journal of computer vision}, 9(2):137--154, 1992.

\bibitem{single_view_mpi}
Richard Tucker and Noah Snavely.
\newblock Single-view view synthesis with multiplane images.
\newblock In {\em Proc. CVPR}, 2020.

\bibitem{tulsiani2017multi}
Shubham Tulsiani, Tinghui Zhou, Alexei~A Efros, and Jitendra Malik.
\newblock Multi-view supervision for single-view reconstruction via
  differentiable ray consistency.
\newblock In {\em Proc. CVPR}, 2017.

\bibitem{tulyakov2018mocogan}
Sergey Tulyakov, Ming-Yu Liu, Xiaodong Yang, and Jan Kautz.
\newblock Mocogan: Decomposing motion and content for video generation.
\newblock In {\em Proceedings of the IEEE conference on computer vision and
  pattern recognition}, pages 1526--1535, 2018.

\bibitem{umeyama1991least}
Shinji Umeyama.
\newblock Least-squares estimation of transformation parameters between two
  point patterns.
\newblock {\em IEEE PAMI}.

\bibitem{Wang15}
Xiaolong Wang and Abhinav Gupta.
\newblock Unsupervised learning of visual representations using videos.
\newblock In {\em Proc. ICCV}, 2015.

\bibitem{Wang_ssganECCV2016}
Xiaolong Wang and Abhinav Gupta.
\newblock Generative image modeling using style and structure adversarial
  networks.
\newblock In {\em ECCV}, 2016.

\bibitem{Wei18}
Donglai Wei, Joseph Lim, Andrew Zisserman, and William~T. Freeman.
\newblock Learning and using the arrow of time.
\newblock In {\em Proc. CVPR}, 2018.

\bibitem{wiles2020synsin}
Olivia Wiles, Georgia Gkioxari, Richard Szeliski, and Justin Johnson.
\newblock Synsin: End-to-end view synthesis from a single image.
\newblock In {\em Proc. CVPR}, 2020.

\bibitem{Wiles18a}
O. Wiles, A.S. Koepke, and A. Zisserman.
\newblock Self-supervised learning of a facial attribute embedding from video.
\newblock In {\em British Machine Vision Conference}, 2018.

\bibitem{xue2016visual}
Tianfan Xue, Jiajun Wu, Katherine Bouman, and Bill Freeman.
\newblock Visual dynamics: Probabilistic future frame synthesis via cross
  convolutional networks.
\newblock In {\em Advances in neural information processing systems}, pages
  91--99, 2016.

\bibitem{YeTian_physicsECCV2018}
Tian Ye, Xiaolong Wang, James Davidson, and Abhinav Gupta.
\newblock Interpretable intuitive physics model.
\newblock In {\em ECCV}, 2018.

\bibitem{IndoorSfMLearner}
Zehao Yu, Lei Jin, and Shenghua Gao.
\newblock P$^{2}$net: Patch-match and plane-regularization for unsupervised
  indoor depth estimation.
\newblock In {\em ECCV}, 2020.

\bibitem{zhang2018unreasonable}
Richard Zhang, Phillip Isola, Alexei~A Efros, Eli Shechtman, and Oliver Wang.
\newblock The unreasonable effectiveness of deep features as a perceptual
  metric.
\newblock In {\em Proceedings of the IEEE conference on computer vision and
  pattern recognition}, pages 586--595, 2018.

\bibitem{zhou2019moving}
Junsheng Zhou, Yuwang Wang, Kaihuai Qin, and Wenjun Zeng.
\newblock Moving indoor: Unsupervised video depth learning in challenging
  environments.
\newblock In {\em Proceedings of the IEEE/CVF International Conference on
  Computer Vision}, pages 8618--8627, 2019.

\bibitem{zhou2017unsupervised}
Tinghui Zhou, Matthew Brown, Noah Snavely, and David~G Lowe.
\newblock Unsupervised learning of depth and ego-motion from video.
\newblock In {\em Proc. CVPR}, 2017.

\bibitem{zhou2018stereo}
Tinghui Zhou, Richard Tucker, John Flynn, Graham Fyffe, and Noah Snavely.
\newblock Stereo magnification: Learning view synthesis using multiplane
  images.
\newblock {\em Proc. ACM SIGGRAPH}, 2018.

\bibitem{zhou2016view}
Tinghui Zhou, Shubham Tulsiani, Weilun Sun, Jitendra Malik, and Alexei~A Efros.
\newblock View synthesis by appearance flow.
\newblock In {\em Proc. ECCV}, 2016.

\bibitem{VON}
Jun-Yan Zhu, Zhoutong Zhang, Chengkai Zhang, Jiajun Wu, Antonio Torralba,
  Joshua~B. Tenenbaum, and William~T. Freeman.
\newblock Visual object networks: Image generation with disentangled 3{D}
  representations.
\newblock In {\em Advances in Neural Information Processing Systems (NeurIPS)},
  2018.

\end{thebibliography}
}
\newpage

\appendix

\twocolumn[
\begin{center}
\section*{\Large Supplementary Material}
\vspace{2em}
\end{center}
]
\section{Additional View Synthesis Results}
\subsection{Our Qualitative Results}
\begin{figure}[b]
    \centering
    \includegraphics[width=0.95\linewidth]{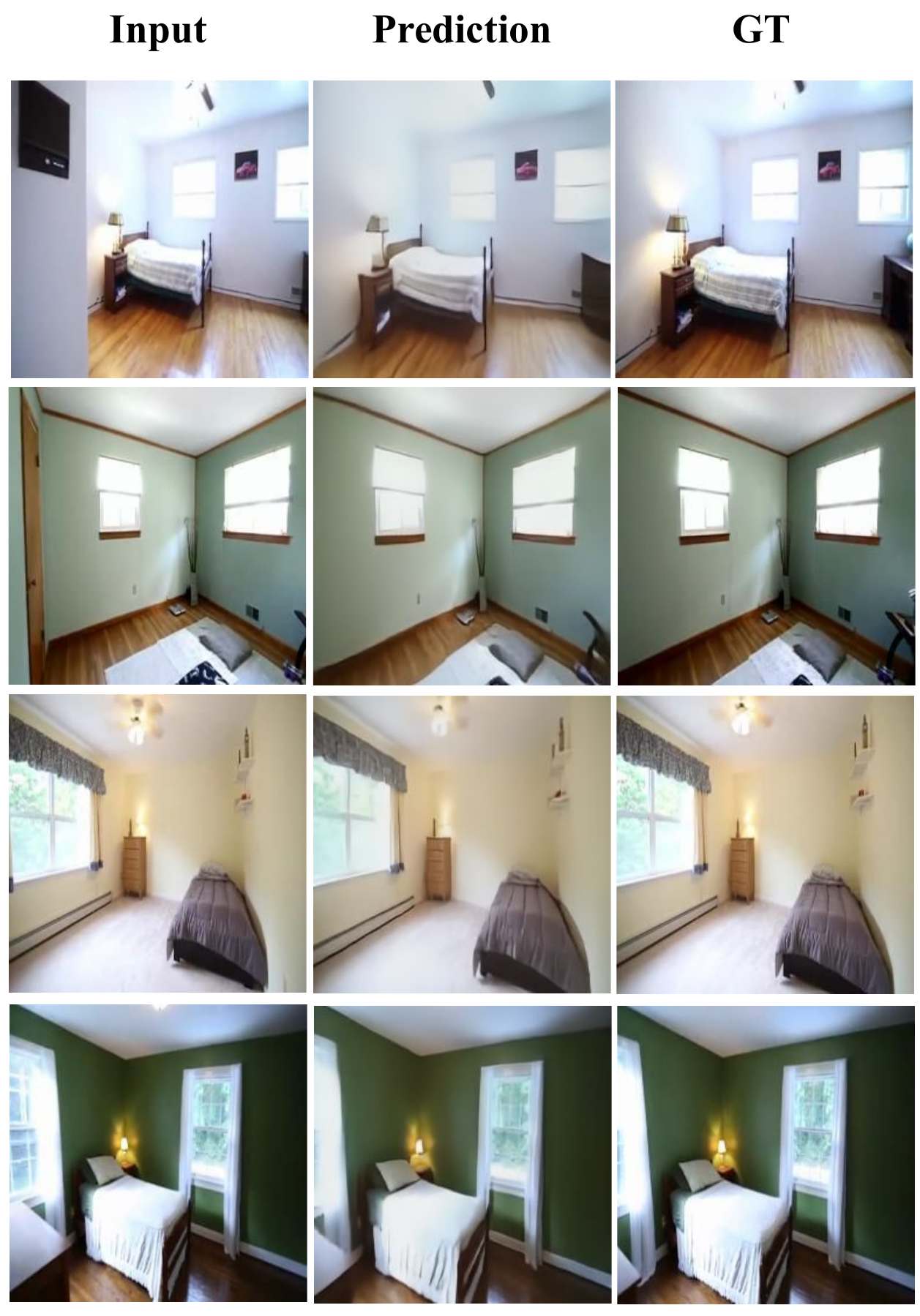}
    \caption{\textbf{Novel View Synthesis:} Our method is able to generate accurate results that correspond to the ground truths. Our method works for both camera translation (row 1-2) and rotation (row 3-4).  }
    \label{fig:qualitative1}
\end{figure}

\begin{figure}[b]
    \centering
    \vspace{-0.1in}
    \includegraphics[width=0.95\linewidth]{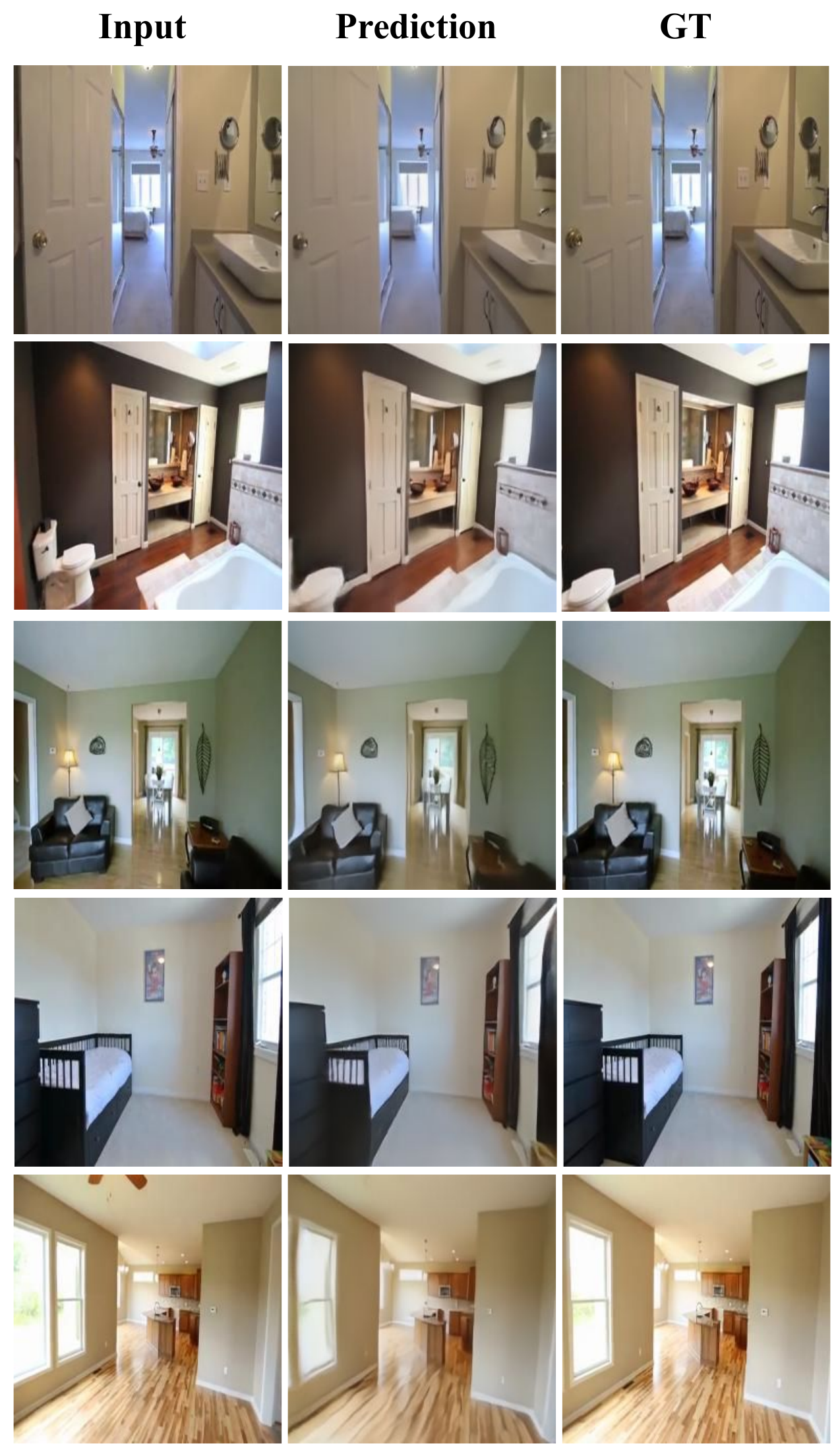}
    \caption{\textbf{Novel View Synthesis (continued)} }
    \label{fig:qualitative2}
\end{figure}

\begin{figure*}[t]
    \centering
    \includegraphics[width=0.9\textwidth]{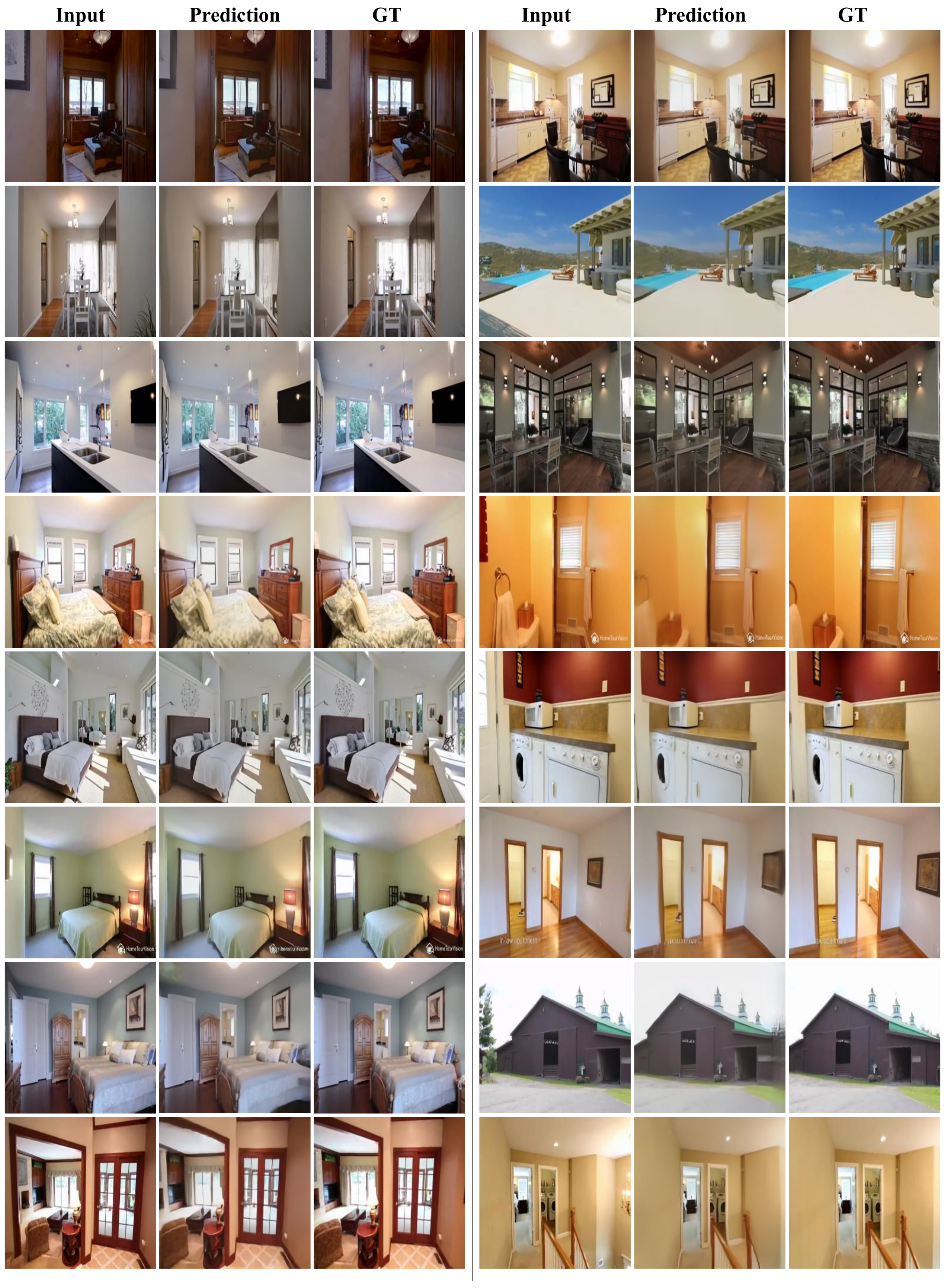}
    \caption{\textbf{Novel View Synthesis (continued)} }
    \label{fig:qualitative3}
\end{figure*}

Figure~\ref{fig:qualitative1}, \ref{fig:qualitative2} and \ref{fig:qualitative3} provide additional novel view synthesis results. Our method is able to synthesize high quality view with correct camera transformation.

\subsection{Qualitative Comparison with Previous Methods}
\begin{figure*}[t]
    \centering
    \includegraphics[width=\textwidth]{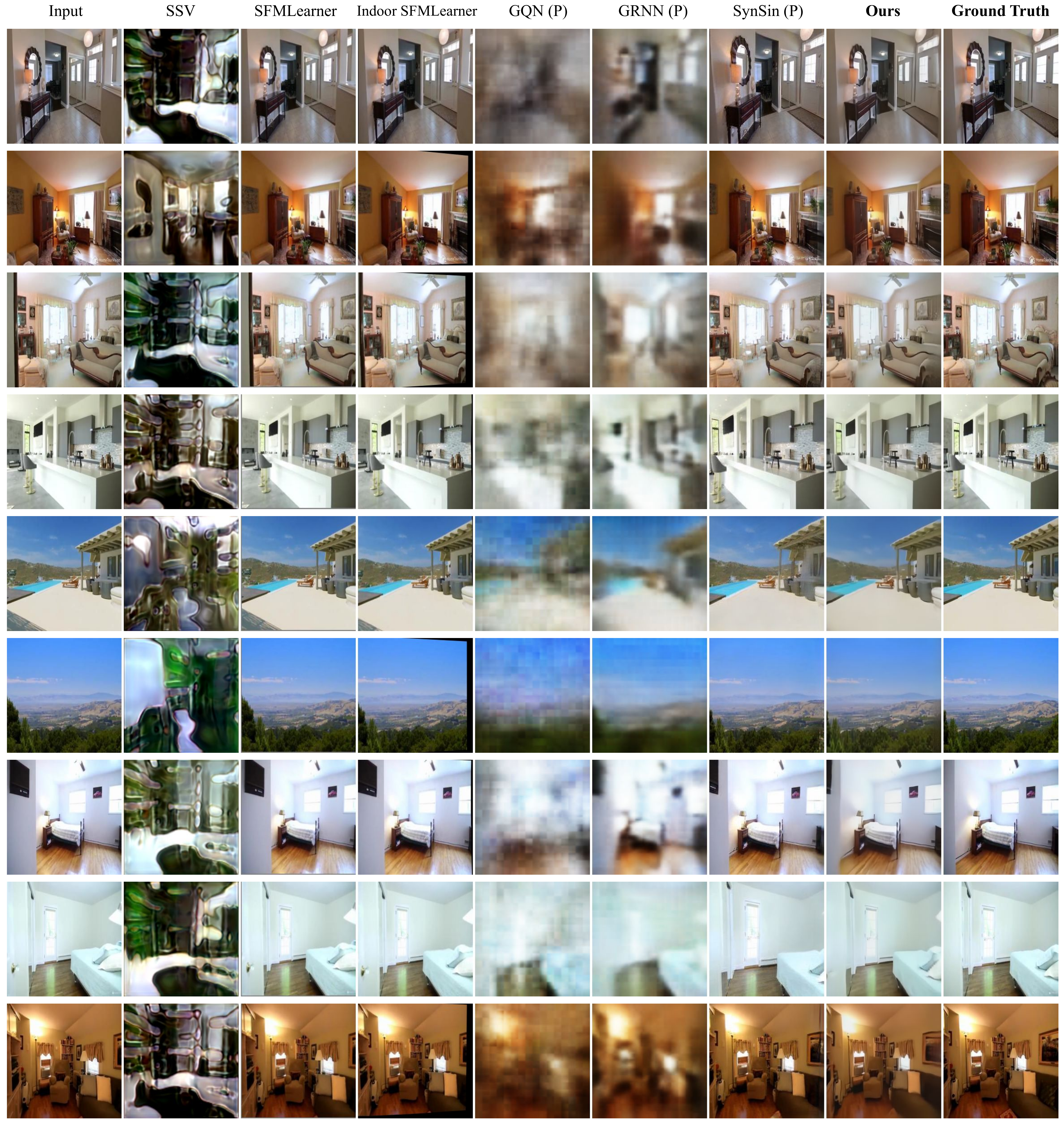}
    \caption{\textbf{Novel view synthesis results compared with previous methods: } Other methods show systematic errors either in rendering or pose estimation. Our method is able to outperform other unsupervised methods visually by large margin. (P) indicates methods supervised by true camera transformations. }
    \vspace{0.5in}
    \label{fig:qualitative_other1}
\end{figure*}

\begin{figure*}[t]
    \centering
    \includegraphics[width=\textwidth]{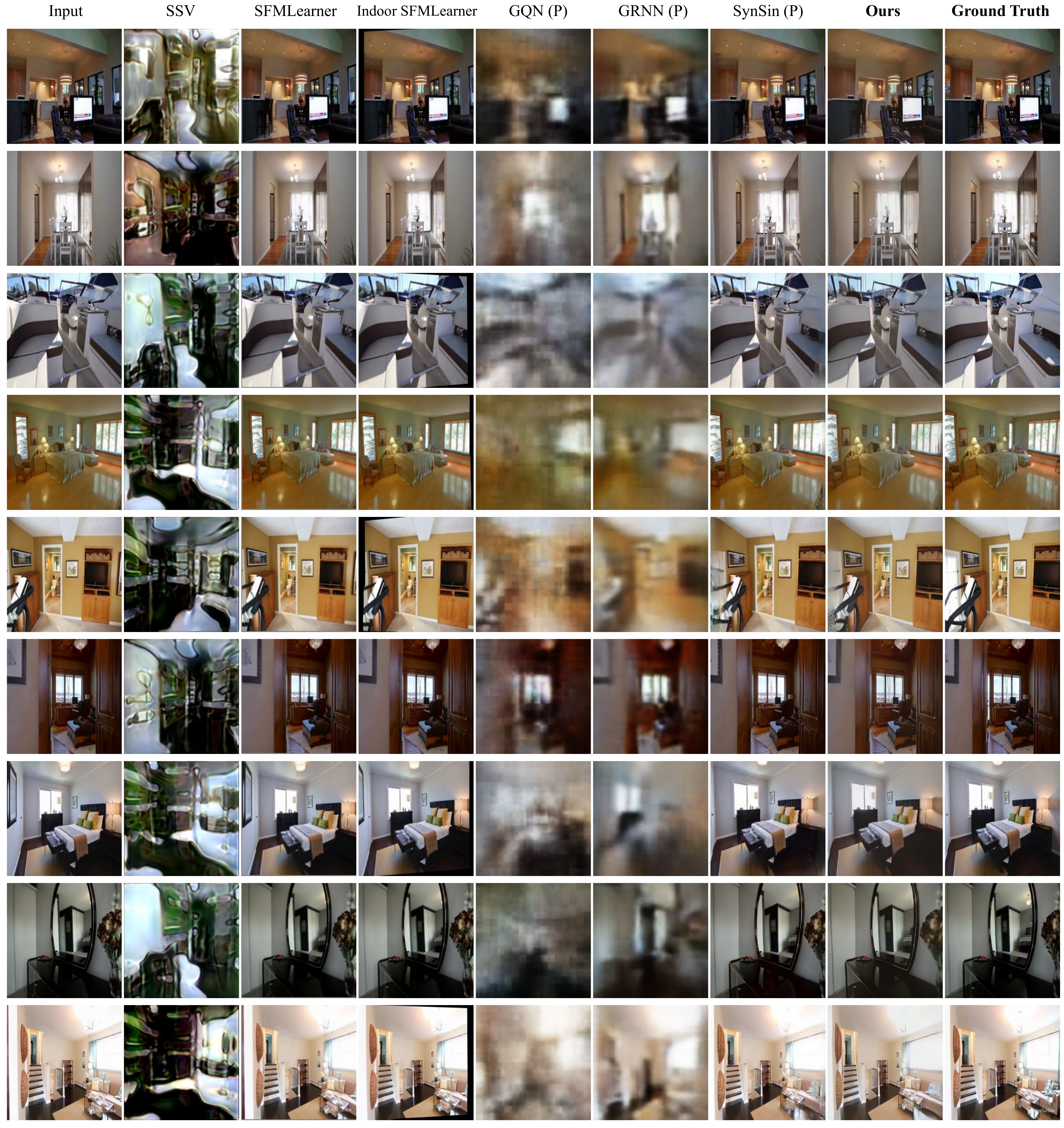}
    \caption{\textbf{Novel view synthesis results compared with previous methods (continued)} }
    \vspace{0.8in}
    \label{fig:qualitative_other2}
\end{figure*}

\begin{figure}[t]
    \centering
    \includegraphics[width=0.73\linewidth]{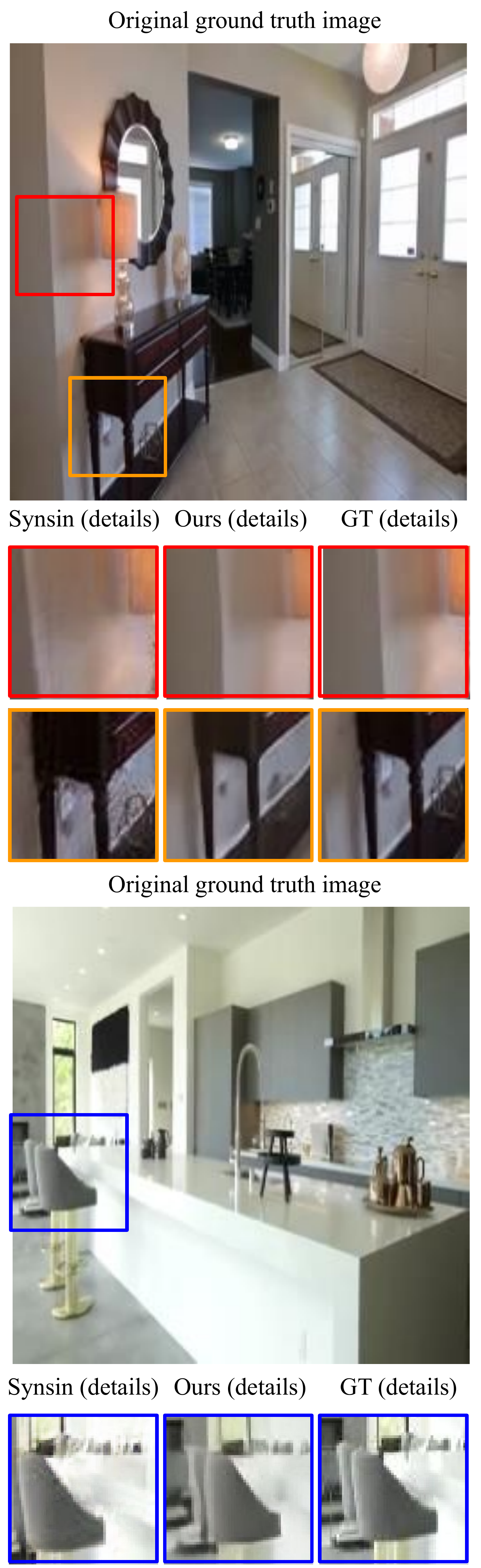}
    \vspace{-0.05in}
    \caption{\textbf{Novel view synthesis results compared with \cite{wiles2020synsin} } Our method synthesizes novel views with less artifacts and noises. See Fig.~\ref{fig:qualitative_other1}  for uncropped full results of our model and \cite{wiles2020synsin}.}
    \label{fig:qualitative_details}
\end{figure}

Figure~\ref{fig:qualitative_other1} and \ref{fig:qualitative_other2}  provide additional comparison with previous methods. In Fig.~\ref{fig:qualitative_details}, we show additional detailed comparison with the strong competitor, Synsin~\cite{wiles2020synsin}. Our method is able to generate more clear and accurate results, even though our method is trained without any camera supervision. 

\section{Network structures}
In Table~\ref{table:network-encoder},\ref{table:network-convnet},\ref{table:network-decoder}, we give additional information about the network strucutres of the subcomponents of our model. There is a total of 29M trainable parameters, mainly in the Resnet-50 feature extractor.

\section{Details of baselines}
In this section, we describe the baselines we used to compare our method with. Specifically, we compared with Dosovitsky et al.~\cite{dosovitskiy2015learning}, Appearance Flow~\cite{zhou2016view}, StereoMag~\cite{zhou2018stereo}, Synsin~\cite{wiles2020synsin}, SFMLearner~\cite{zhou2017unsupervised}, Indoor SFMLearner~\cite{IndoorSfMLearner}, GQN~\cite{eslami2018neural}, GRNN~\cite{Tung_2019_CVPR} and SSV~\cite{mustikovela2020self}.
\newpage

\textbf{Dosovitsky et al.} This method infers a novel view of a given input with a neural network. The model feeds an image and the desired viewpoint into the network, which directly estimates an image corresponding to the given view. This method was only tested on images of a single object, which are considerably easier than the real scene data we used.

\textbf{Appearance Flow.} Appearance flow predicts a flow field that warps the original image into a novel view. This flow field is computed from a ConvNet which takes the original image and the viewpoint transformation as input. This method is shown to work very well on synthetic object datasets (\textit{e.g.} ShapeNet~\cite{shapenet2015}). It has also shown that the performance on real-world scenes surpasses direct pixel prediction~\cite{tatarchenko2016multi}.

\textbf{SynSin.}
Synsin is a novel view synthesis algorithm that aims to synthesize new viewpoints of a given single image. The algorithm makes use of an intermediate representation: a point cloud where each point is a feature vector. Taking a single image as input, Synsin first computes a depth map and a set of 2D feature maps. The 2D feature maps are projected back into 3D points with the depth map. Next, a camera transformation is given and the point cloud is rendered with the new camera pose. There are two major differences in the high-level philosophy between our model and Synsin. First, we do not require true camera transformation during training time. This means that Video Autoencoder works on raw videos and Synsin can only be trained on datasets with camera pose obtained from sensors or precomputed from SfM. Second, Synsin leverages explicit 3D representation (\textit{i.e.} the depth map) whereas the Video Autoencoder does not. As shown in the paper, explicit 3D representation could fail in out-of-domain data whereas our method generalizes better into unseen images.

\textbf{StereoMag.} StereoMag constructs multiplane images (MPIs) from two input views. The multiplane images represent a 3D structure as a set of fronto-parallel planes at fixed depths. It can be rendered at a given viewpoint by \textit{blending} planes with a set of blending weights. These blending weights are predicted from one image, and the other image is used as the plane sweep volume (\textit{i.e.} the value of the planes). Apart from the representation, one major difference between our method and StereoMag is that StereoMag makes use of two images as input, which greatly simplifies the problem. 

\textbf{SFMLearner.} SFMLearner disentangle a short video clip into depth maps and camera trajectory. The method is structurally similar to our method. However, there are several major differences. (i) SFMLearner (and similarly, GeoNet, etc.) results were shown on the relatively simple KITTI dataset, but work poorly on more complex data~\cite{zhou2019moving}. (ii) SFMLearner requires ground truth camera intrinsics, which makes it more difficult to train on raw videos. (iii) SFMLearner cannot produce satisfactory view synthesis results because its spatial representation is a 2.5D depth map.  We use the predicted depth and camera transformation to warp the first frame into the target frame. 
\vspace{5em}
\begin{table}[t]
\centering
\footnotesize\addtolength{\tabcolsep}{2pt}
\setlength{\extrarowheight}{2pt}

\begin{tabular}{ccc}

 \toprule
 Stage & Configuration & Output \\
  \hline
 0  & Input image & $H\times W \times 3$ \\
 \hline
  \multicolumn{3}{c} { \textbf{2D feature extraction}} \\
 \hline
1 & Extract feature with Resnet-50 & $\frac{H}{16}\times \frac{W}{16} \times 2048$ \\
\hline
  \multicolumn{3}{c} { \textbf{Reshaping 2D to 3D}} \\
 \hline
2 & Reshape feature dimension to 256 & $\frac{H}{16}\times \frac{W}{16} \times 8 \times 256$ \\ 
\hline
  \multicolumn{3}{c} { \textbf{3D Convolutions}} \\
 \hline
3 &  \makecell{ 3D-Deconvolution \\  ($4^3$ kernel, 128 filters, stride 2) } & $\frac{H}{8}\times \frac{W}{8} \times 16 \times 128$ \\ 
4 & \makecell{ 3D-Deconvolution \\  ($4^3$ kernel, 32 filters, stride 2) } & $\frac{H}{4}\times \frac{W}{4} \times 32 \times 32$ \\ 
\bottomrule

\end{tabular}
\caption{3D Encoder ($\mathcal{F}_{3D}$) architecture. 
\label{table:network-encoder}}

\end{table}

\begin{table}[t]
\centering
\footnotesize\addtolength{\tabcolsep}{2pt}
\setlength{\extrarowheight}{2pt}

\begin{tabular}{ccc}

 \toprule
 Stage & Configuration & Output \\
  \hline
 0  & Two concatenated input image & $H\times W \times 6$ \\
 \hline
  \multicolumn{3}{c} { \textbf{2D feature extraction}} \\
 \hline
1 &  \makecell{ 2D-convolution \\  ($3^2$ kernel, 16 filters, stride 2) } & $\frac{H}{2}\times \frac{W}{2} \times 16$ \\ 
2 &  \makecell{ 2D-convolution \\  ($3^2$ kernel, 32 filters, stride 2) } & $\frac{H}{4}\times \frac{W}{4} \times 32$ \\ 
3 &  \makecell{ 2D-convolution \\  ($3^2$ kernel, 64 filters, stride 2) } & $\frac{H}{8}\times \frac{W}{8} \times 64$ \\ 
4 &  \makecell{ 2D-convolution \\  ($3^2$ kernel, 128 filters, stride 2) } & $\frac{H}{16}\times \frac{W}{16} \times 128$ \\ 
5 &  \makecell{ 2D-convolution \\  ($3^2$ kernel, 256 filters, stride 2) } & $\frac{H}{32}\times \frac{W}{32} \times 256$ \\ 
6 &  \makecell{ 2D-convolution \\  ($3^2$ kernel, 256 filters, stride 2) } & $\frac{H}{64}\times \frac{W}{64} \times 256$ \\ 
7 &  \makecell{ 2D-convolution \\  ($3^2$ kernel, 256 filters, stride 2) } & $\frac{H}{128}\times \frac{W}{128} \times 256$ \\ 
8 &  \makecell{ 2D-convolution \\  ($1^2$ kernel, 6 filters, stride 1) } & 
$\frac{H}{128}\times \frac{W}{128} \times 6$ \\ 
9 & Mean pooling & $6$ \\ 
10 & Multiply by 0.01 & $6$ \\ 
\bottomrule

\end{tabular}
\caption{Architecture of ConvNet ($\mathcal{H}$) for Trajectory Encoder ($\mathcal{F}_{Traj}$). The last step could stabilize the training process.
\label{table:network-convnet}}
\end{table}

\begin{table}[t]
\centering
\footnotesize\addtolength{\tabcolsep}{2pt}
\setlength{\extrarowheight}{2pt}

\begin{tabular}{ccc}

 \toprule
 Stage & Configuration & Output \\
  \hline
 0  & Input 3D deep voxels & $\frac{H}{4}\times \frac{W}{4} \times 32 \times 32$ \\
 1 & Input 3D transformation & $6$ \\
 \hline
  \multicolumn{3}{c} { \textbf{Rotating 3D deep voxels}} \\
 \hline

 2  & Rotate deep voxels with input transf. & $\frac{H}{4}\times \frac{W}{4} \times 32 \times 32$ \\
3 &  \makecell{ 3D-Convolution \\  ($3^3$ kernel, 64 filters, stride 1) } & $\frac{H}{4}\times \frac{W}{4} \times 32 \times 64$ \\ 
4 & \makecell{ 3D-Convolution \\  ($3^3$ kernel, 64 filters, stride 1) } & $\frac{H}{4}\times \frac{W}{4} \times 32 \times 64$ \\ 
 \hline
  \multicolumn{3}{c} { \textbf{Reshape 3D to 2D}} \\
 \hline
5 & Concatenate feature and depth dim. &  $\frac{H}{4}\times \frac{W}{4} \times 2048$ \\
\hline
  \multicolumn{3}{c} { \textbf{2D Convolutions (neural network renderer)}} \\
 \hline
6 &  \makecell{ 2D-convolution \\  ($1^2$ kernel, 512 filters, stride 1) } & $\frac{H}{4}\times \frac{W}{4} \times 512$ \\ 
7 & \makecell{ 2D-Deconvolution \\  ($4^2$ kernel, 64 filters, stride 2) } & $\frac{H}{2}\times \frac{W}{2} \times 64$ \\ 
8 & \makecell{ 2D-Deconvolution \\  ($4^2$ kernel, 32 filters, stride 2) } & $H\times W \times 32$ \\ 
9 & \makecell{ 2D-Deconvolution \\  ($3^2$ kernel, 3 filters, stride 1) } & $H\times W \times 3$ \\ 
\bottomrule

\end{tabular}
\caption{Decoder ($\mathcal{G}$) architecture. 
\label{table:network-decoder}}

\end{table}

\textbf{Indoor SFMLearner.} P$^2$-Net, or the Indoor SFMLearner, proposes to use a patch-based loss to overcome the optimization problem of SFMLearner, and shows better results in indoor scenes. Similar to SFMLearner, we use predicted depth and camera to warp the first frame into the target frame. Because not all pixels in the target frame have a corresponding pixel in the first frame, this method could produce large blank areas (as seen in Fig.~\ref{fig:qualitative_other1} and \ref{fig:qualitative_other2}).

\textbf{GQN} Generative Query Network combines 2D features of multiple input images into a 2D \textit{Neural Scene Representation}. This 2D representation is then decoded with an LSTM network conditioned on a query viewpoint. Although GQN is shown to yield good results on toy datasets, we find it struggles to generate clear results for real-world scenes.

\textbf{GRNN} GRNN constructs an RNN-aggregated 3D voxel from input images as scene representation. This 3D voxel could be projected back into a set of 2D feature maps with a specific query viewpoint and decoded into a query view. While GRNN also makes use of a 3D deep voxel as representation, several significant differences between our model and GRNN include: (i) GRNN only support 2 degrees of freedom for camera transformations (yaw and pitch) whereas we support a full 6 DOF camera transformation (ii) GRNN is trained with ground truth cameras whereas we do not make use camera supervision. (iii) GRNN requires a complicated matching process during testing. As shown in the main text and Fig.~\ref{fig:qualitative_other1} and \ref{fig:qualitative_other2}, GRNN fails to produce satisfactory results on RealEstate10K.

\textbf{SSV.} SSV is the state-of-the-art self-supervised viewpoint estimation algorithm. Leveraging multiple constraints such as cycle consistencies, SSV learns from image collections to predict camera poses of these images. We retrained SSV on the same training dataset, treating video frames as individual images. The output of SSV is a rotation angle of a single image. To obtain the relative pose, we concatenate pose predictions of the reference frame and current frame. We then fit a linear regression model to predict the true camera transformation using about 600 true poses, similar to SSV's original testing procedure. For fair comparisons, we also applied the Umeyama alignment~\cite{umeyama1991least}.

\end{document}